\documentclass{article}

\usepackage[margin=1in,footskip=0.25in]{geometry}

\usepackage[utf8]{inputenc} 
\usepackage[T1]{fontenc}    
\usepackage{url}            
\usepackage{booktabs}       
\usepackage{amsfonts}       
\usepackage{nicefrac}       
\usepackage{microtype}      

\usepackage{subfigure}
\usepackage{subfiles}
\usepackage{graphicx}
\usepackage{stackengine}
\usepackage{amsmath,amsfonts,bm}
\usepackage{caption}
\usepackage{multirow}
\usepackage{array}

\usepackage{amssymb}
\usepackage{pifont}
\newcommand{\cmark}{\ding{51}}%
\newcommand{\xmark}{\ding{55}}%

\usepackage{wrapfig}
\usepackage[]{authblk}

\title{Batch Normalization is a Cause of Adversarial Vulnerability}

\author[1,2]{Angus Galloway}
\author[2,3,4]{Anna Golubeva}
\author[5]{Thomas Tanay\thanks{Work done while at University College
London.}}
\author[1]{Medhat Moussa}
\author[1,2,6]{Graham W.~Taylor}
\affil[1]{School of Engineering, University of Guelph}
\affil[2]{Vector Institute for Artificial Intelligence}
\affil[3]{Department of Physics and Astronomy, University of Waterloo}
\affil[4]{Perimeter Institute for Theoretical Physics}
\affil[5]{Huawei Noah’s Ark Lab}
\affil[6]{Google Brain}

\def\biblio{\bibliographystyle{abbrv}\bibliography{../icml2019_zotero}}

\begin{document}
\def\biblio{}

\date{\vspace{-5ex}}
\maketitle


\begin{abstract}
Batch normalization (batch norm) is often used in an attempt to stabilize and
accelerate training in deep neural networks. In many cases it indeed decreases the
number of parameter updates required to achieve low training error. However,
it also reduces robustness to small adversarial input perturbations and noise by
double-digit percentages, as we show on five standard datasets.
Furthermore, substituting weight decay for batch norm is sufficient to nullify
the
relationship between adversarial vulnerability and the input dimension.
Our work is consistent with a mean-field analysis that found that batch norm
causes exploding gradients.

\end{abstract}

\section{Introduction}
\label{sec:introduction}

Batch norm is a standard component of modern deep neural
networks, and tends to make the training process less sensitive to the
choice of hyperparameters in many cases~\cite{ioffe2015batch}.
While ease of training is desirable for model developers, an important concern
among stakeholders is that of model robustness to plausible, previously unseen inputs
during deployment.

The adversarial examples phenomenon has exposed unstable predictions across
state-of-the-art models~\cite{szegedy2014intriguinga}. This has
led to a variety of methods that aim to improve robustness, but
doing so effectively remains a challenge~\cite{athalye2018obfuscated,
schott2018towards, hendrycks2019benchmarking, jacobsen2019exploiting}.
We believe that a prerequisite to developing methods that increase
robustness is an understanding of factors that reduce it.

\begin{wrapfigure}{r}{0.45\columnwidth} 
\begin{center}
\includegraphics[width=0.43\columnwidth]{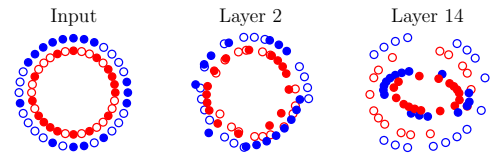}
\end{center}
\caption{Two mini-batches from the ``Adversarial Spheres'' dataset (2D),
and their representations in a deep linear network with batch norm at initialization.
Mini-batch membership is indicated by marker fill and class membership
by colour. Each layer is projected to its first two principal components.
Classes are mixed by Layer 14.}
\label{fig:adv-circle}
\end{wrapfigure}

Approaches for improving robustness often begin with existing neural network
architectures---that use batch norm---and patching them against specific
attacks, e.g., through inclusion of adversarial examples during
training~\cite{szegedy2014intriguinga, goodfellow2015explaining,
kurakin2017adversarial, madry2018towards}. An implicit assumption is that batch
norm itself does not reduce robustness -- an assumption that we tested empirically 
and found to be invalid.
In the original work that introduced batch norm, it was suggested that other
forms of regularization can be turned down or disabled when using it without
decreasing
standard test accuracy. Robustness, however, is less forgiving: it is strongly
impacted by the disparate mechanisms of various regularizers.

The frequently made observation that adversarial vulnerability can scale with the
input dimension~\cite{goodfellow2015explaining, gilmer2018adversarial,
simon-gabriel2018adversarial}
highlights the importance of identifying regularizers as more than
merely a way to improve test accuracy.
In particular, batch norm was a confounding factor in~\cite{simon-gabriel2018adversarial},
making the results of their initialization-time analysis hold after training.
By adding $\ell_2$ regularization and removing batch norm, we show that there
is no~\emph{inherent} relationship between adversarial vulnerability and the
input dimension.



\biblio


\setlength{\abovedisplayskip}{5pt} 
\setlength{\belowdisplayskip}{5pt}

\section{Batch Normalization}
\label{sec:background}

We briefly review how batch norm modifies the hidden layers' pre-activations $h$
of a neural network. We use the notation
of~\cite{yang2019mean}, where $\alpha$ is the index for a neuron, $l$
for the layer, and $i$ for a mini-batch of $B$ samples from the dataset;
$N_l$ denotes the number of neurons in layer $l$, $W^l$ is the matrix of weights
and $b^l$ is the vector of biases that parametrize layer $l$.
The batch mean is defined as $\mu_\alpha = \frac{1}{B} \sum_i h_{\alpha
i}$, and the variance is $\sigma_\alpha^2 = \frac{1}{B}
\sum_i {(h_{\alpha i} - \mu_\alpha)}^2$.
In the batch norm procedure, the mean $\mu_\alpha$ is subtracted from the
pre-activation of each unit $h_{\alpha i}^l$ (consistent
with~\cite{ioffe2015batch}),
the result is divided by the standard deviation $\sigma_\alpha$ plus a small
constant $c$ to prevent division by zero, then scaled and shifted
by the learned parameters $\gamma_\alpha$ and $\beta_\alpha$, respectively.
This is described in Eq.~\eqref{eq:bn}, where a per-unit nonlinearity $\phi$,
e.g., ReLU, is applied after the normalization.
\begin{equation} \label{eq:bn}
h_i^l = W^l \phi (\tilde{h}_i^{l - 1}) + b^l, \quad \quad \tilde{h}_{\alpha i}^l =
\gamma_{\alpha} \frac{h_{\alpha i} - \mu_{\alpha}}{\sqrt{\sigma_\alpha^2 + c}} +
\beta_\alpha
\end{equation}

Note that this procedure fixes the first and second moments of all neurons
$\alpha$ equally at initialization, independent of the width or depth of the network.
This suppresses the information contained in these moments. Because batch norm
induces a non-local batch-wise nonlinearity to each unit $\alpha$, this
loss of information cannot be recovered by the parameters $\gamma_\alpha$
and $\beta_\alpha$. Furthermore, it has been widely observed empirically that these
parameters do not influence the effect being studied~\cite{vanlaarhoven2017l2, zhang2019three, yang2019mean}.
Thus, $\gamma_\alpha$ and $\beta_\alpha$ can be incorporated into the
per-unit nonlinearity without loss of generality.

To understand how batch normalization is harmful, consider two mini-batches that
differ by only a~\emph{single} example: due to the induced batch-wise nonlinearity,
they will have different representations for~\emph{each} example~\cite{yang2019mean}.
This difference is further amplified by stacking batch norm layers.
Conversely, normalization of intermediate representations for two different inputs
impair the ability of batch-normalized networks to distinguish high-quality examples
(as judged by an ``oracle'') that ought to be classified with a large prediction margin,
from low-quality, i.e., more ambiguous, instances.
The last layer of a discriminative neural network, in particular, is typically a
linear decoding of class label-homogeneous clusters, and thus makes extensive use
of information represented via differences in mean and variance at this stage
for the purpose of classification.

We argue that this information loss and inability to maintain relative
distances in the input space reduces adversarial as well as general robustness.
Figure~\ref{fig:adv-circle} shows a degradation of class-relevant input
distances in a batch-normalized linear network on a 2D variant of the
``Adversarial Spheres''
dataset~\cite{gilmer2018adversarial}.\footnote{We add a ReLU nonlinearity when
attempting to~\emph{learn} the binary classification task posed
by~\cite{gilmer2018adversarial}. In Appendix~\ref{sec:spheres}
we show that batch norm increases sensitivity to the learning rate in this case.}
Conversely, class membership is preserved in arbitrarily deep unnormalized
networks (See Figure~\ref{fig:sp-no-bn} of Appendix~\ref{sec:spheres}),
but we require a scaling factor to increase the magnitude of the
activations to see this visually.

\biblio

\section{Empirical Result}
\label{sec:empirical}

We first evaluate the robustness (quantified as the drop in test accuracy under
input perturbations)
of convolutional networks, with and without batch norm, that were trained using
standard
procedures. The datasets -- MNIST, SVHN, CIFAR-10, and ImageNet -- 
were normalized to zero mean and unit variance.

As a white-box adversarial attack we use projected gradient descent (PGD),
$\ell_\infty$- and $\ell_2$-norm variants, for its
simplicity and ability to degrade performance
with little perceptible change to the input~\cite{madry2018towards}.
We run PGD for 20 iterations, with $\epsilon_\infty = 0.03$ and a step size of
$\epsilon_\infty / 10$ for SVHN, CIFAR-10, and $\epsilon_\infty = 0.01$ for ImageNet.
For PGD-$\ell_2$ we set $\epsilon_2 = \epsilon_\infty \sqrt{d}$, where $d$ is
the input dimension.
We report the test accuracy for additive Gaussian noise of zero mean and
variance $\nicefrac{1}{4}$, denoted as ``Noise''~\cite{ford2019adversarial}, as
well as the CIFAR-10-C common corruption
benchmark~\cite{hendrycks2019benchmarking}.
We found these methods were sufficient
to demonstrate a considerable disparity in robustness due to batch norm, but
this is not intended as a formal security evaluation. All uncertainties are the
standard error of the mean.\footnote{Each experiment has a unique uncertainty,
hence the number of decimal places varies.}


\begin{wraptable}{R}{9cm}
\caption{Test accuracies of VGG8 on SVHN.}
\begin{tabular}{ccccc} \toprule
BN & Clean & Noise & PGD-$\ell_\infty$ & PGD-$\ell_2$ \\ \midrule
\xmark & $92.60 \pm 0.04$ & $83.6 \pm 0.2$ & $27.1 \pm 0.3$ & $22.0 \pm 0.8$ \\ \midrule
\cmark & $94.46 \pm 0.02$ & $78.1 \pm 0.6$ & $10 \pm 1$ & $1.6 \pm 0.3$ \\ \bottomrule
\end{tabular}
\label{tab:svhn}
\end{wraptable}

For the SVHN dataset, models were trained by stochastic gradient descent (SGD)
with momentum 0.9 for 50 epochs,
with a batch size of 128 and initial learning rate of $0.01$, which was dropped by
a factor of ten at epochs 25 and 40. Trials were repeated over five random seeds.
We show the results of this experiment in Table~\ref{tab:svhn}, finding that
despite batch norm increasing clean test accuracy by $1.86 \pm 0.05\%$, it reduced
test accuracy for additive noise by $5.5 \pm 0.6\%$, for PGD-$\ell_\infty$ by
$17 \pm 1\%$, and for PGD-$\ell_2$ by $20 \pm 1\%$.


\begin{table}[h]
\centering
\caption{Test accuracies of VGG8 and WideResNet--28--10 on CIFAR-10 and
CIFAR-10.1 (\texttt{v6}) in several variants: clean, noisy, and PGD perturbed.}
\begin{tabular}{cccccccc}
\toprule
\multicolumn{2}{c}{} & \multicolumn{4}{c}{CIFAR-10} &
\multicolumn{2}{c}{CIFAR-10.1} \\ \midrule
Model & BN & Clean & Noise & PGD-$\ell_\infty$ & PGD-$\ell_2$ & Clean & Noise \\ \midrule
VGG & \xmark & $87.9 \pm 0.1$ & $79 \pm 1$ & $52.9 \pm 0.6$ & $65.6 \pm 0.3$
& $75.3 \pm 0.2$ & $66 \pm 1$ \\ \midrule
VGG & \cmark & $88.7 \pm 0.1$ & $73 \pm 1$ & $35.7 \pm 0.3$ & $59.7 \pm 0.3$
& $77.3 \pm 0.2$ & $60 \pm 2$ \\ \midrule
WRN & F & $94.6 \pm 0.1$ & $69 \pm 1$ & $20.3 \pm 0.3$ & $9.4 \pm 0.2$
& $87.5 \pm 0.3$ & $68 \pm 1$\\ \midrule
WRN & \cmark & $95.9 \pm 0.1$ & $58 \pm 2$ & $14.9 \pm 0.6$ & $8.3 \pm 0.3$
& $89.6 \pm 0.2$ & $58 \pm 1$ \\ \bottomrule
\end{tabular}
\label{tab:cifar10-and-10p1}
\end{table}

For the CIFAR-10 experiments we trained models with a similar procedure as for
SVHN, but with random $32 \times 32$ crops using four-pixel
padding, and horizontal flips.
We evaluate two families of contemporary models, one without skip connections
(VGG), and WideResNets (WRN) using ``Fixup'' initialization~\cite{zhang2019residual}
to reduce the use of batch norm.

In the first experiment, a basic comparison with and without batch norm shown in
Table~\ref{tab:cifar10-and-10p1}, we evaluated the best model in terms of test
accuracy
after training for 150 epochs with a fixed learning rate of $0.01$.
In this case, inclusion of batch norm for VGG reduces the clean generalization gap
(difference between training and test accuracy) by $1.1 \pm 0.2\%$. For
additive noise, test accuracy drops by $6 \pm 1\%$, and for PGD perturbations
by $17.3 \pm 0.7\%$ and $5.9 \pm 0.4\%$ for $\ell_\infty$ and $\ell_2$ variants,
respectively.

\begin{wraptable}{R}{7.5cm}
\caption{VGG models of increasing depth on CIFAR-10, with and without batch norm
(BN). See text for differences in hyperparameters compared to Table~\ref{tab:cifar10-and-10p1}.}
\begin{tabular}{ccccc}
\hline
\multicolumn{2}{c}{Model} & \multicolumn{3}{c}{Test Accuracy ($\%$)} \\[0.05cm] \hline
L & BN & Clean & Noise & PGD-$\ell_\infty$ \\[0.05cm] \hline
8 & \xmark & $89.29 \pm 0.09$ & $81.7 \pm 0.3$ & $55.6 \pm 0.4$ \\ \hline
8 & \cmark & $90.49 \pm 0.01$ & $77 \pm 1$ & $40.6 \pm 0.6$ \\ \hline \hline
11 & \xmark & $90.4 \pm 0.1$ & $81.5 \pm 0.5$ & $53.7 \pm 0.2$ \\ \hline
11 & \cmark & $91.19 \pm 0.06$ & $79.3 \pm 0.6$ & $43.8 \pm 0.5$ \\ \hline \hline
13 & \xmark & $91.74 \pm 0.02$ & $77.8 \pm 0.7$ & $40.3 \pm 0.7$ \\ \hline
13 & \cmark & $93.0 \pm 0.1$ & $67 \pm 1$ & $28.5 \pm 0.4$ \\ \hline
16 & \cmark & $92.8 \pm 0.1$ & $66 \pm 2$ & $28.9 \pm 0.2$ \\ \hline
19 & \cmark & $92.65 \pm 0.09$ & $68 \pm 2$ & $30.0 \pm 0.1$ \\ \hline
\end{tabular}
\label{tab:cifar10-opt-acc}
\end{wraptable}

Very similar results are obtained on a new test set,
CIFAR-10.1~\texttt{v6}~\cite{recht2018cifar10a}:
batch norm slightly improves the clean test
accuracy (by $2.0 \pm 0.3\%$), but leads to a considerable drop in test accuracy
of $6 \pm 1\%$ for the case with additive noise, and
$15 \pm 1\%$ and $3.4 \pm 0.6\%$ respectively for $\ell_\infty$ and $\ell_2$ PGD
variants (PGD absolute values omitted for CIFAR-10.1 in
Table~\ref{tab:cifar10-and-10p1} for brevity).

It has been suggested that one of the benefits of batch norm is that it
facilitates training with a larger learning rate~\cite{ioffe2015batch,
bjorck2018understanding}. We test this from a robustness perspective in an
experiment summarized in
Table~\ref{tab:cifar10-opt-acc}, where the initial learning rate was increased
to $0.1$ when batch norm was used. We prolonged training for up to 350 epochs, and
dropped the learning rate by a factor of ten at epoch 150 and 250 in both cases,
which increases clean test accuracy relative to Table~\ref{tab:cifar10-and-10p1}.
The deepest model that was trainable using standard ``He''
initialization~\cite{he2015delving}
without batch norm was VGG13.~\footnote{For which one of ten random seeds failed
to achieve better than chance accuracy on the training set, while others
performed as expected. We report the first three successful runs for consistency
with the other experiments.} None of the deeper batch-normalized models recovered
the robustness of the most shallow, or same-depth unnormalized equivalents,
nor does the higher learning rate with batch norm improve
robustness compared to baselines trained for the same number of
epochs. Additional results for deeper models on SVHN and CIFAR-10 can be found
in Appendix~\ref{sec:appendix-deeper}.


We also evaluated robustness on the common corruption benchmark comprising 19 types
of real-world effects that can be grouped
into four categories: ``noise'', ``blur'', ``weather'',
and ``digital'' corruptions~\cite{hendrycks2019benchmarking}.
Each corruption has five ``severity'' or intensity levels.
We report the mean error on the corrupted test set (mCE) by averaging over all
intensity levels and corruptions~\cite{hendrycks2019benchmarking}.
We summarize the results for two VGG variants and a WideResNet on
CIFAR-10-C, trained from scratch on the default training set for three and five
random seeds respectively.
Accuracy for the noise corruptions, which caused the largest difference in
accuracy with batch norm, are outlined in Table~\ref{tab:cifar10-c}.

The key takeaway is:~\emph{For all models tested, the batch-normalized variant
had a higher error rate for all corruptions of the ``noise'' category, at every
intensity level}.

\begin{table}[]
\centering
\caption{Robustness of three modern convolutional neural network
architectures with and without batch norm on the
\texttt{CIFAR-10-C} common ``noise'' corruptions~\cite{hendrycks2019benchmarking}.
We use ``F'' to denote the Fixup variant of WRN. Values were averaged over five
intensity levels for each corruption.}
\begin{tabular}{ccccccc}
\toprule
\multicolumn{2}{c}{Model} & \multicolumn{5}{c}{Test Accuracy ($\%$)} \\ \midrule
Variant & BN & Clean & Gaussian & Impulse & Shot & Speckle \\ \midrule
\multirow{2}{*}{VGG8} & \xmark & $87.9 \pm 0.1$ & $\mathbf{65.6 \pm 1.2}$
 & $\mathbf{58.8 \pm 0.8}$ & $\mathbf{71.0 \pm 1.2}$ & $\mathbf{70.8 \pm 1.2}$ \\ \cline{2-7}
 & \cmark & $88.7\pm 0.1$ & $56.4 \pm 1.5$ & $51.2 \pm 0.1$ & $65.4 \pm 1.1$
 & $66.3 \pm 1.1$ \\ \midrule 
\multirow{2}{*}{VGG13} & \xmark & $91.74 \pm 0.02$ & $\mathbf{64.5\pm 0.8}$
 & $\mathbf{63.3 \pm 0.3}$ & $\mathbf{70.9 \pm 0.4}$ & $\mathbf{71.5 \pm 0.5}$ \\ \cline{2-7}
 & \cmark & $93.0 \pm 0.1$ & $43.6 \pm 1.2$ & $49.7 \pm 0.5$ & $56.8 \pm 0.9$
 & $60.4 \pm 0.7$ \\ \midrule 
\multirow{2}{*}{WRN28} & F & $94.6 \pm 0.1$ & $\mathbf{63.3 \pm 0.9}$
 & $\mathbf{66.7 \pm 0.9}$ & $\mathbf{71.7 \pm 0.7}$ & $\mathbf{73.5 \pm 0.6}$ \\ \cline{2-7}
 & \cmark & $95.9 \pm 0.1$ & $51.2 \pm 2.7$ & $56.0 \pm 2.7$ & $63.0 \pm 2.5$
 & $66.6 \pm 2.5$ \\ \bottomrule
\end{tabular}
\label{tab:cifar10-c}
\end{table}

Averaging over all 19 corruptions we find that batch norm increased mCE by
$1.9 \pm 0.9\%$ for VGG8, $2.0 \pm 0.3\%$ for VGG13, and $1.6 \pm 0.4\%$ for
WRN\@. There was a large disparity in accuracy when modulating batch norm
for different corruption categories, therefore we examine these in more detail.

\begin{wraptable}{r}{7cm}
\caption{Models from~\texttt{torchvision.models} pre-trained on ImageNet,
some with and some without batch norm (BN).}
\begin{tabular}{ccccc}
\hline
\multicolumn{2}{c}{Model} & \multicolumn{3}{c}{Top 5 Test Accuracy ($\%$)} \\[0.05cm] \hline
Model & BN & Clean & Noise & PGD-$\ell_\infty$ \\[0.05cm] \hline
VGG-11 & \xmark & $88.63$ & $49.16$ & $37.12$ \\ \hline
VGG-11 & \cmark & $89.81$ & $49.95$ & $26.12$ \\ \hline
VGG-13 & \xmark & $89.25$ & $52.55$ & $29.16$ \\ \hline
VGG-13 & \cmark & $90.37$ & $52.12$ & $20.63$ \\ \hline
VGG-16 & \xmark & $90.38$ & $60.67$ & $32.81$ \\ \hline
VGG-16 & \cmark & $91.52$ & $65.36$ & $21.96$ \\ \hline
VGG-19 & \xmark & $90.88$ & $64.86$ & $34.19$ \\ \hline
VGG-19 & \cmark & $91.84$ & $68.79$ & $24.49$ \\ \hline
AlexNet & \xmark & $79.07$ & $41.41$ & $39.12$ \\ \hline
DenseNet121 & \cmark & $91.97$ & $79.85$ & $34.75$ \\ \hline
ResNet18 & \cmark & $88.65$ & $79.62$ & $31.07$ \\ \hline
\end{tabular}
\label{tab:imagenet}
\end{wraptable}

For VGG8, the mean generalization gaps for noise were:
Gaussian---$9.2 \pm 1.9\%$, Impulse---$7.5 \pm 0.8\%$,
Shot---$5.6 \pm 1.6\%$, and Speckle---$4.5 \pm 1.6\%$.
After the ``noise'' category the next most damaging corruptions (by
difference in accuracy due to batch norm) were:
Contrast---$4.4 \pm 1.3\%$, Spatter---$2.4 \pm 0.7\%$,
JPEG---$2.0 \pm 0.4\%$, and Pixelate---$1.3 \pm 0.5\%$. Results for the remaining
corruptions were a coin toss as to whether batch norm improved or degraded
robustness, as the random error was in the same ballpark as the difference being
measured.
For VGG13, the batch norm accuracy gap enlarged to $26-28\%$ for Gaussian noise
at severity levels 3, 4, and 5; and over $17\%$ for Impulse noise at levels 4
and 5.
Averaging over all levels, we have gaps for noise variants of:
Gaussian---$20.9 \pm 1.4\%$, Impulse---$13.6 \pm 0.6\%$,
Shot---$14.1 \pm 1.0\%$, and Speckle---$11.1 \pm 0.8\%$.
Robustness to the other
corruptions seemed to benefit from the slightly higher clean test accuracy
of $1.3 \pm 0.1\%$ for the batch-normalized VGG13. The remaining
generalization gaps varied from (negative) $0.2 \pm 1.3\%$ for Zoom blur, to
$2.9 \pm 0.6\%$ for Pixelate.

For the WRN, the mean generalization gaps for noise were:
Gaussian---$12.1 \pm 2.8\%$, Impulse---$10.7 \pm 2.9\%$,
Shot---$8.7 \pm 2.6\%$, and Speckle---$6.9 \pm 2.6\%$. Note that the large
uncertainty for these measurements is due to high variance for the model
with batch norm, on average $2.3\%$ versus $0.7\%$ for Fixup. JPEG compression
was next at $4.6 \pm 0.3\%$.
Interestingly, some corruptions that led to a
positive gap for VGG8 showed a negative gap for the
WRN, i.e.,~batch norm improved accuracy to:
Contrast---$4.9 \pm 1.1\%$, Snow---$2.8 \pm 0.4\%$, Spatter---$2.3 \pm 0.8\%$.
These were the same corruptions for which VGG13 lost, or did not improve its
robustness when batch norm was removed, hence why we believe these correlate
with standard test accuracy (highest for WRN). Visually, these
corruptions appear to preserve texture information. Conversely,
noise is applied in a spatially global way that disproportionately degrades
these textures, emphasizing shapes and edges.
It is now known that modern CNNs trained on standard image datasets have a
propensity to rely on texture, but we would rather they use shape and edge
cues~\cite{geirhos2019imagenettrained, brendel2019approximating}. Our results
support the idea that batch norm may be exacerbating this tendency to leverage
superficial textures for classification of image data.



Next, we evaluated the robustness of pre-trained ImageNet models from
the~\texttt{torchvision.models} repository, which conveniently provides models
with and without batch norm.\footnote{\url{https://pytorch.org/docs/stable/torchvision/models.html}, \texttt{v1.1.0}.}
Results are shown in~Table~\ref{tab:imagenet}, where batch norm improves top-5
accuracy on noise in some cases, but consistently reduces it by $8.54\%$ to
$11.00\%$ (absolute) for PGD. The trends are the same for top-1
accuracy, only the absolute values were smaller; the degradation varies from
$2.38\%$ to $4.17\%$. Given the discrepancy between noise and
PGD for ImageNet, we include a black-box transfer analysis in
Appendix~\ref{sec:imagenet-black-box} that is consistent with the white-box analysis.


\begin{figure}
\centering
\includegraphics[width=.8\columnwidth]{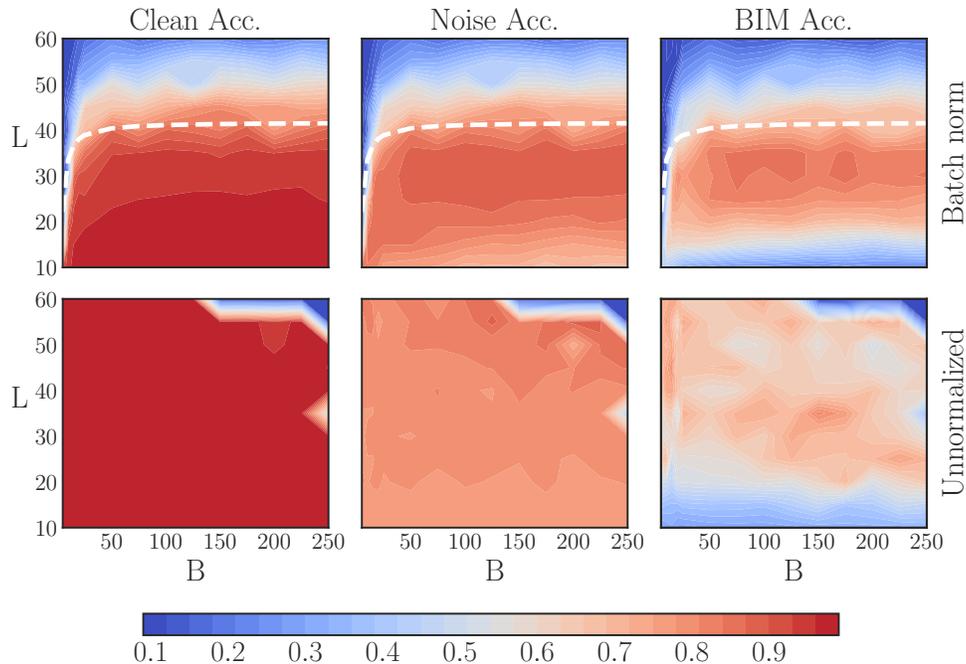}
\caption{We extend the experiment of~\cite{yang2019mean} by training
fully-connected nets of depth $L$ and constant-width ($N_l=384$) ReLU layers by
SGD, batch size $B$, and learning rate $\eta = 10^{-5} B$ on MNIST\@.
The batch norm parameters $\gamma$ and $\beta$ were left as default, momentum
disabled, and $c = 10^{-3}$. The dashed line is the theoretical maximum
trainable depth of batch-normalized networks as a function of the batch size.
We report the clean test accuracy, and that for additive Gaussian noise and BIM
perturbations.
The batch-normalized models were trained for 10 epochs, while the unnormalized
were trained for 40 epochs as they took longer to converge. The 40 epoch
batch-normalized plot was qualitatively similar with dark blue bands for BIM for
shallow and deep variants.
The dark blue patch for 55 and 60 layer unnormalized models at large batch sizes
depicts a total failure to train. These networks were trainable by reducing
$\eta$, but for consistency we keep $\eta$ the same in both cases.}
\label{fig:mft-batch}
\end{figure}

Finally, we explore the role of batch size and depth in
Figure~\ref{fig:mft-batch}. Batch norm limits the maximum trainable
depth, which~\emph{increases} with the batch size, but quickly plateaus as
predicted by Theorem 3.10 of~\cite{yang2019mean}.
Robustness~\emph{decreases} with the batch size
for depths that maintain a reasonable test accuracy, at around 25 or fewer layers.
This tension between clean accuracy and robustness as a function of the batch size
is not observed in unnormalized networks.

\begin{wrapfigure}{R}{0.5\columnwidth} 
\includegraphics[width=.48\columnwidth]{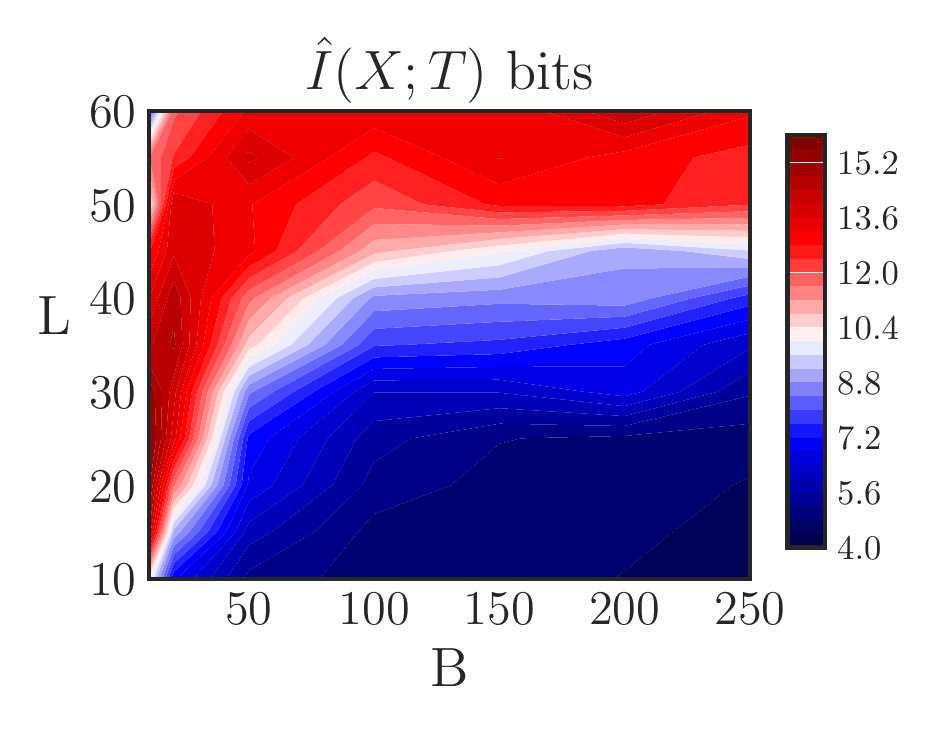}
\caption{Estimated mutual information $\hat{I}(X; T)$ between quantized
representations $T$ and input $X$ for batch-normalized
models from Figure~\ref{fig:mft-batch}. Values are lower-bounded by
$\log_2 (10) \approx 3.32$ by the number of classes and upper-bounded by
$\log_2(60000) \approx 15.87$ by the number of samples in the training set.
Estimates accurate to within $\pm 0.32$ bits of $I(X; T)$~\cite{paninski2003estimationa}.}
\label{fig:mft-ixt}
\end{wrapfigure}

In unnormalized networks, we observe that perturbation robustness increases with
the depth of the network. This is
consistent with the computational benefit of the hidden layers proposed
by~\cite{shwartz-ziv2017opening}, who take an information-theoretic approach.
This analysis uses two mutual information terms: $I(X; T)$ -- the information in
the layer activations $T$
about the input $X$, which is a measure of representational complexity,
and $I(T; Y)$ -- the information in the activations $T$ about the label $Y$,
which is understood as the predictive content of the learned input representations $T$.
It is shown that under SGD training, $I(T; Y)$ generally increases with the
number of epochs, while $I(X; T)$ increases initially, but reduces throughout the
later stage of the training procedure. An information-theoretic proof
as to why reducing $I(X; T)$, while ensuring a sufficiently high value of $I(T; Y)$,
should promote good generalization from finite samples is given in~\cite{tishby2015deepb, shwartz-ziv2019representation}.

We estimate $\hat{I}(X; T)$ of the batch-normalized networks from
the experiment in Figure~\ref{fig:mft-batch} for sub-sampled batch sizes and
plot it in Figure~\ref{fig:mft-ixt}.
We assume $I(X; T) = H(T) - H(T | X) = H(T)$, since the networks are
noiseless and thus $T$ is deterministic given $X$. We use the ``plug-in''
maximum-likelihood estimate of the entropy $H(T)$ using the full MNIST training
set~\cite{paninski2003estimationa}. Activations $T$ are taken as the softmax
output, which was quantized to 7-bit accuracy. The number of bits was determined by
reducing the precision as low as possible without inducing classification errors.
This provides a notion of the model's ``intrinsic'' precision.
We use the confidence interval: $\log_2 \big(1 + (m - 1)/ N \big) + N^{-1/2}\log_2(m)$
recommended by~\cite{paninski2003estimationa}, which contains both bias and
variance terms for $\hat{H}_{MLE}$ for the regime
$N \gg m$ where $N=60000$ and $m=2^7$. This is multiplied by ten for each
dimension.

Our first observation is that the configurations where
$\hat{I}(X; T)$ is low---indicating a more compressed representation---are
the same settings where the model obtains high clean test accuracy.
The transition of $\hat{I}(X; T)$ at around 10 bits occurs remarkably close to
the theoretical maximum trainable depth of $\approx 40$ layers.
For the unnormalized network, the absolute values of $\hat{I}(X; T)$ were almost
always in the same ballpark or less than the lowest value obtained by any
batch-normalized network, which was $4.26 \pm 0.32$ bits.
We therefore omit the
comparable figure for brevity, but note that $\hat{I}(X; T)$ did continue to
decrease with depth in many cases, e.g., from $4.21$ $(L=10)$ to $4.06$ $(L=60)$ for
a mini-batch size of 20, but unfortunately these differences were small compared to
the worst-case error. The fact that $\hat{I}(X; T)$ is small where BIM robustness
is poor for batch-normalized networks disagrees with our initial hypothesis
that more layers were needed to decrease $\hat{I}(X; T)$. However, this result is
consistent with the observation that it is possible for networks to overfit via
too much compression~\cite{shwartz-ziv2017opening}.
In particular,~\cite{yang2019mean} prove that batch norm loses the information
between mini-batches exponentially quickly in the depth of the network, so
over-fitting via ``too much'' compression is consistent with our results. This intuition
requires further analysis, which is left for future work.

\biblio

\section{Vulnerability and Input Dimension}
\label{sec:input-dimension}

A recent work~\cite{simon-gabriel2018adversarial} analyzes adversarial vulnerability of batch-normalized
networks at initialization time and conjectures based on a scaling analysis that,
under the commonly used~\cite{he2015delving} initialization scheme,
adversarial vulnerability scales as $\sim\sqrt{d}$.

\begin{wraptable}{r}{7.5cm}
\caption{Evaluating the robustness of a MLP with and without batch norm.
See text for architecture. We observe a $61 \pm 1\%$ reduction in
test accuracy due to batch norm for $\sqrt{d}=84$ compared to $\sqrt{d}=28$.}
\begin{tabular}{ccccr} 
\toprule
\multicolumn{2}{c}{Model} & \multicolumn{3}{c}{Test Accuracy ($\%$)} \\ \midrule
$\sqrt{d}$ & BN & Clean & Noise & $\epsilon = 0.1$ \\ \midrule
\multirow{2}{*}{28} & \xmark & $97.95 \pm 0.08$ & $93.0 \pm 0.4$ & $66.7 \pm 0.9$ \\ \cline{2-5}
 & \cmark & $97.88 \pm 0.09$ & $76.6 \pm 0.7$ & $22.9 \pm 0.7$ \\ \hline 
\multirow{2}{*}{56} & \xmark & $98.19 \pm 0.04$ & $93.8 \pm 0.1$ & $53.2 \pm 0.7$ \\ \cline{2-5}
 & \cmark & $98.22 \pm 0.02$ & $79.3 \pm 0.6$ & $8.6 \pm 0.8$ \\ \hline 
\multirow{2}{*}{84} & \xmark & $98.27 \pm 0.04$ & $94.3 \pm 0.1$ & $47.6 \pm 0.8$ \\ \cline{2-5}
 & \cmark & $98.28 \pm 0.05$ & $80.5 \pm 0.6$ & $6.1 \pm 0.5$ \\ \bottomrule
\end{tabular}
\label{tab:three-layer-bn}
\end{wraptable}

They also show in experiments that independence between
vulnerability and the input dimension can be approximately recovered through adversarial
training by projected gradient descent (PGD)~\cite{madry2018towards}, with a
modest trade-off of clean accuracy. We show that this can be achieved by simpler
means and with little to no trade-off through $\ell_2$ weight decay, where the
regularization constant $\lambda$ corrects the loss scaling as the norm
of the input increases with $d$. 

\begin{wraptable}{r}{7.5cm}
\caption{Evaluating the robustness of a MLP with $\ell_2$ weight decay
(same $\lambda$ as for linear model, see Table 5 of Appendix~\ref{sec:appendix-input-dimension}). See text for architecture.
Adding batch norm degrades all accuracies.}
\begin{tabular}{ccccr} 
\toprule
\multicolumn{2}{c}{Model} & \multicolumn{3}{c}{Test Accuracy ($\%$)} \\ \midrule
$\sqrt{d}$ & BN & Clean & Noise & $\epsilon = 0.1$ \\ \midrule 
\multirow{2}{*}{56} & \xmark & $97.62 \pm 0.06$ & $95.93 \pm 0.06$ & $87.9 \pm 0.2$  \\ \cline{2-5}
 & \cmark & $96.23 \pm 0.03$ & $90.22 \pm 0.18$ & $66.2 \pm 0.8$ \\ \hline 
\multirow{2}{*}{84} & \xmark & $96.99 \pm 0.05$ & $95.69 \pm 0.09$ & $87.9 \pm 0.1$ \\ \cline{2-5}
 & \cmark & $93.30 \pm 0.09$ & $87.72 \pm 0.11$ & $65.1 \pm 0.5$ \\ \bottomrule
\end{tabular}
\label{tab:three-layer-wd}
\end{wraptable}

We increase the MNIST image width $\sqrt{d}$ from 28
to 56, 84, and 112 pixels. The loss $\mathcal{L}$ is predicted to grow like
$\sqrt{d}$ for $\epsilon$-sized attacks by Thm.~4
of~\cite{simon-gabriel2018adversarial}. We confirm
that without regularization the loss does scale roughly as predicted:
the predicted values lie between loss ratios obtained for
$\epsilon = 0.05$ and $\epsilon = 0.1$ attacks for most image widths
(see Table 4 of Appendix~\ref{sec:appendix-input-dimension}).
Training with $\ell_2$ weight decay, however, we obtain adversarial test
accuracy ratios of $0.98 \pm 0.01$, $0.96 \pm 0.04$, and $1.00 \pm 0.03$ and
clean accuracy ratios of $0.999 \pm 0.002$, $0.996 \pm 0.003$, and $0.987 \pm
0.004$ for $\sqrt{d}$ of 56, 84, and 112 respectively, relative to the
original $\sqrt{d} = 28$ dataset. A more detailed explanation and results are
provided in Appendix~\ref{sec:appendix-input-dimension}.

Next, we repeat this experiment with a two-hidden-layer ReLU MLP, with the
number of hidden units equal to the half the input dimension, and
optionally use one hidden layer with batch norm.\footnote{This choice
of architecture is mostly arbitrary, the trends were the same for constant
width layers.}
To evaluate robustness, 100 iterations of BIM-$\ell_\infty$ were used with
a step
size of 1e-3, and $\epsilon_\infty = 0.1$.
We also report test accuracy with additive Gaussian noise of zero mean and
unit variance, the same first two moments as the clean images.\footnote{We first
apply the noise to the original 28$\times$28 pixel images, then resize them to
preserve the appearance of the noise.}

Despite a difference in clean accuracy of only $0.08 \pm 0.05\%$,
Table~\ref{tab:three-layer-bn} shows that for the original image resolution,
batch norm reduced accuracy for noise by $16.4 \pm 0.4\%$, and for
BIM-$\ell_\infty$ by $43.8 \pm 0.5\%$. Robustness keeps decreasing as the
image size increases, with the batch-normalized network having
$\sim 40\%$ less robustness to BIM and $13-16\%$ less to noise at all sizes.

We then apply the $\ell_2$ regularization constants tuned for the respective
input dimensions on the linear model to the ReLU MLP with no further adjustments.
Table~\ref{tab:three-layer-wd} shows that by adding sufficient $\ell_2$
regularization ($\lambda = 0.01$) to recover the original ($\sqrt{d} = 28$, no BN)
accuracy for BIM of $\approx 66 \%$ when using batch norm,
we induce a test error increase of $1.69 \pm 0.01\%$, which
is substantial on MNIST\@. Furthermore, using the same regularization constant
without batch norm increases clean test accuracy by $1.39 \pm 0.04\%$, and
for the BIM-$\ell_\infty$ perturbation by $21.7 \pm 0.4\%$.

Following the guidance in the original work on batch norm~\cite{ioffe2015batch} to the
extreme ($\lambda = 0$): to~\emph{reduce} weight decay when using batch norm,
accuracy for the $\epsilon_\infty = 0.1$ perturbation is degraded by
$79.3 \pm 0.3\%$ for $\sqrt{d} = 56$, and $81.2 \pm 0.2\%$ for $\sqrt{d} = 84$.

\emph{In all cases, using batch norm greatly reduced test accuracy for noisy and
adversarially perturbed inputs, while weight decay increased accuracy for such inputs.}

\biblio

\section{Related Work}
\label{sec:related}

Our work examines the effect of batch norm on model robustness at test time.
Many references which have an immediate connection to our work were discussed
in the previous sections; here we briefly mention other works that do not have
a direct relationship to our experiments, but are relevant to the topic of batch
norm in general.

The original work~\cite{ioffe2015batch} that introduced batch norm as a technique for improving neural
network training and test performance motivated it by the ``internal covariate shift'' --
a term refering to the changing distribution of layer outputs, an effect that requires subsequent
layers to steadily adapt to the new distribution and thus slows down the training process.
Several follow-up works started from the empirical observation that batch norm usually
accelerates and stabilizes training, and attempted to clarify the mechanism behind this effect.
One argument is that batch-normalized networks have a smoother optimization
landscape due to smaller gradients immediately before the batch-normalized
layer~\cite{santurkar2018how}. However,~\cite{yang2019mean} study the
effect of stacking many batch-normalized layers and prove that this
causes gradient explosion that is exponential in the depth of the network for any
non-linearity. In practice, relatively shallow batch-normalized
networks yield the expected ``helpful smoothing'' of the loss surface
property~\cite{santurkar2018how}, while very deep networks are not
trainable~\cite{yang2019mean}. In our work, we find that a single
batch-normalized layer suffices to induce severe adversarial vulnerability.

\newcommand{\mnist}{.075}
\begin{wrapfigure}{r}{0.6\columnwidth}
\begin{tabular}{m{0mm}c}
& \multirow{3}{*}{
\subfigure{
\stackunder[3pt]{\includegraphics[width=\mnist\columnwidth]{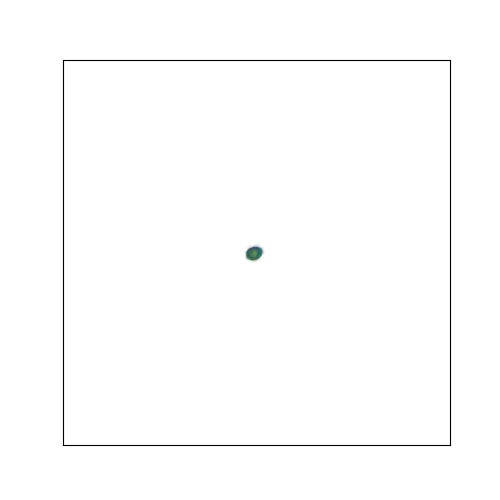}}{}
\stackunder[3pt]{\includegraphics[width=\mnist\columnwidth]{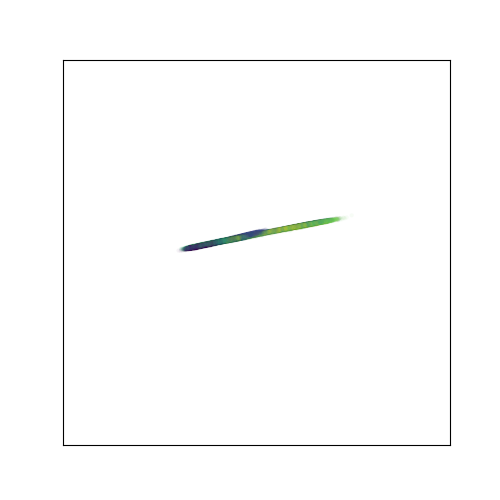}}{}
\stackunder[3pt]{\includegraphics[width=\mnist\columnwidth]{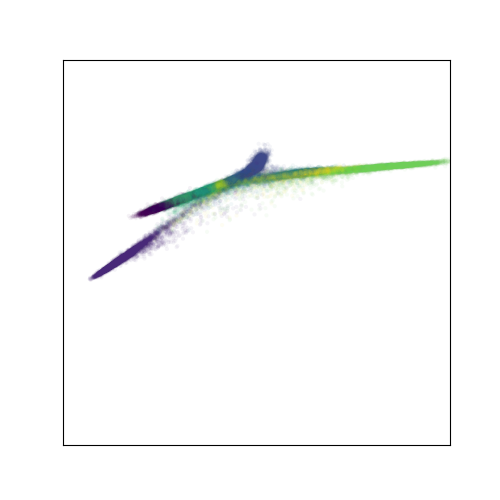}}{}
\stackunder[3pt]{\includegraphics[width=\mnist\columnwidth]{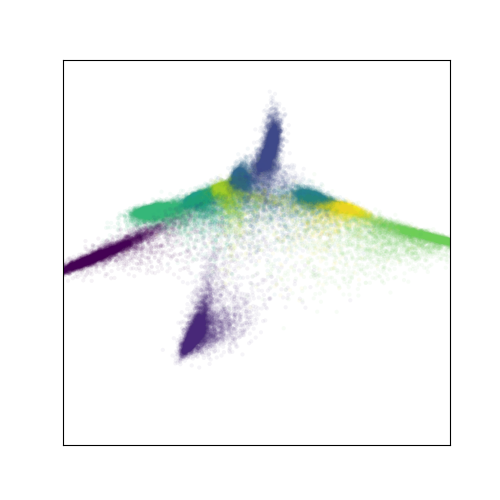}}{}
\stackunder[3pt]{\includegraphics[width=\mnist\columnwidth]{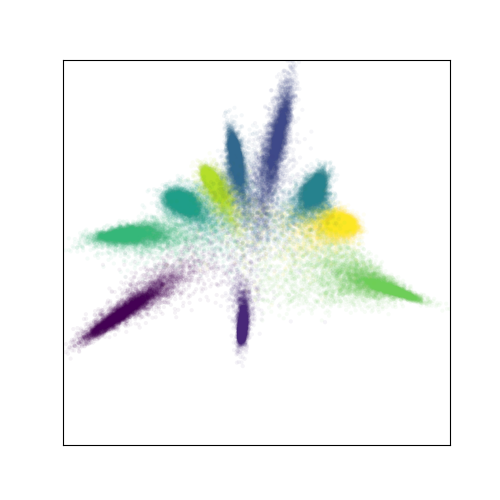}}{}
\stackunder[3pt]{\includegraphics[width=\mnist\columnwidth]{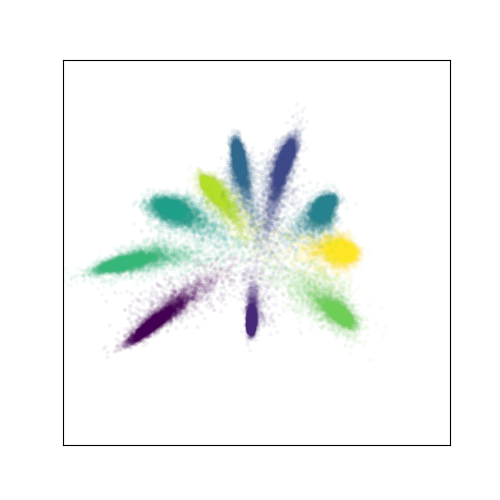}}{}
\stackunder[3pt]{\includegraphics[width=\mnist\columnwidth]{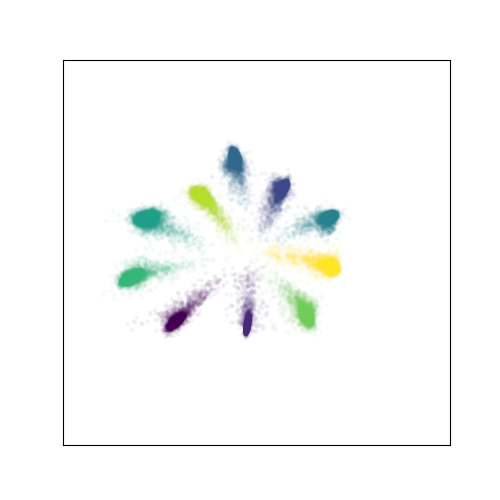}}{}\label{fig:a}}} \\ [3pt] \\
(a) & \\
& \\
& \multirow{3}{*}{
\subfigure{
\stackunder[3pt]{\includegraphics[width=\mnist\columnwidth]{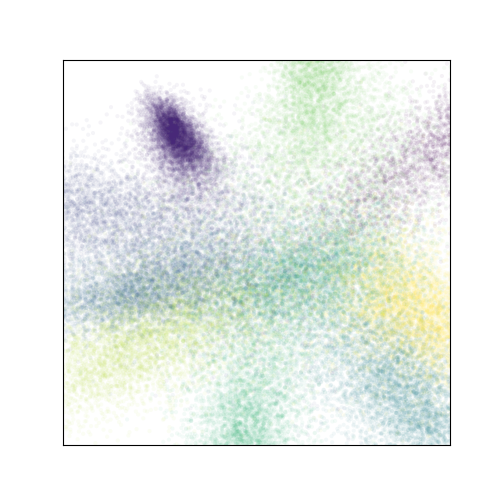}}{}
\stackunder[3pt]{\includegraphics[width=\mnist\columnwidth]{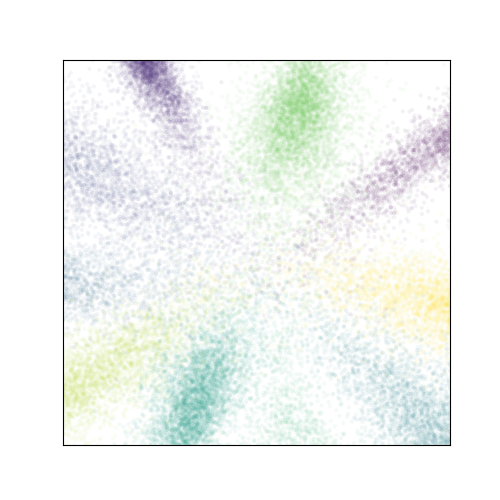}}{}
\stackunder[3pt]{\includegraphics[width=\mnist\columnwidth]{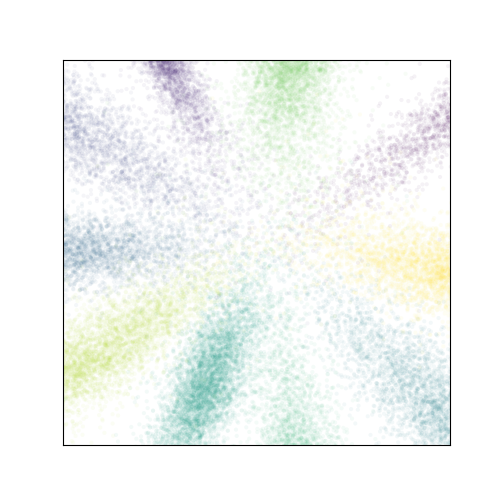}}{}
\stackunder[3pt]{\includegraphics[width=\mnist\columnwidth]{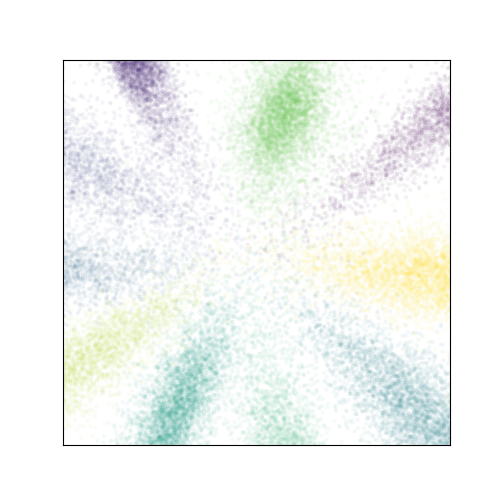}}{}
\stackunder[3pt]{\includegraphics[width=\mnist\columnwidth]{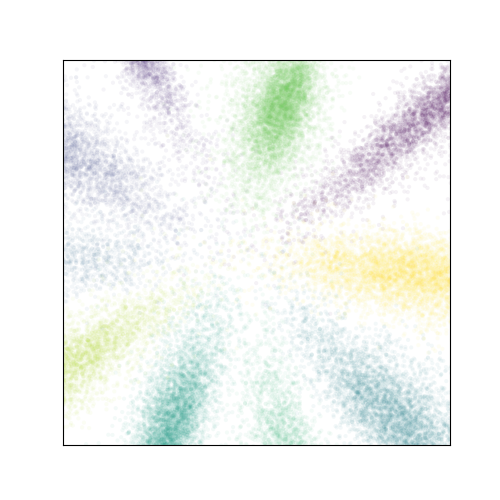}}{}
\stackunder[3pt]{\includegraphics[width=\mnist\columnwidth]{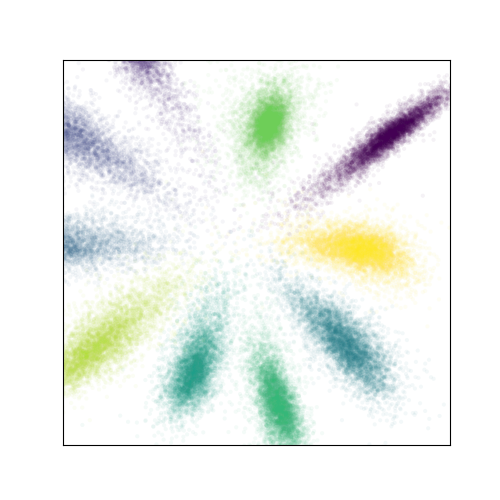}}{}
\stackunder[3pt]{\includegraphics[width=\mnist\columnwidth]{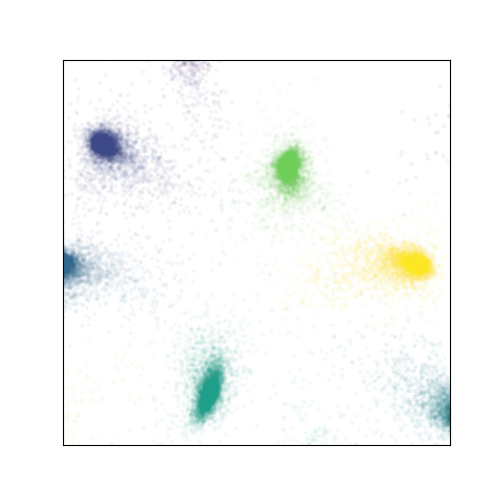}}{}\label{fig:b}}} \\ \\
(b) & \\
&
\end{tabular}
\vspace{5mm}
{\caption{Visualization of activations in a two-unit layer over 500 epochs.
Model is a fully-connected MLP [\subref{fig:a} (784--392--196--2--49--10)
and~\subref{fig:b} (784--392--BN--196--BN--2--49--10)]
with ReLU units, mini-batch size 128, learning rate 1e-2, and weight decay $\lambda$=1e-3.
The plots have a fixed x- and y-axis range of $\pm 10$.
Samples from the MNIST training set are plotted and colored by
label.}
\label{fig:mnist-sp}}
\end{wrapfigure}

In Figure~\ref{fig:mnist-sp} we visualize
the activations of the penultimate hidden layer in a fully-connected
network~\subref{fig:a} without and~\subref{fig:b} with batch norm over the
course of 500 epochs.
In the unnormalized network~\ref{fig:a}, all data points are overlapping at
initialization.
Over the first $\approx20$ epochs, the points spread further apart (middle plot)
and begin to form clusters. In the final stage, the clusters become tighter.

When we introduce two batch-norm layers in the network, placing them before the
visualized layer, the activation patterns display notable differences, as shown
in Figure~\ref{fig:b}:
i) at initialization, all data points are spread out, allowing easier
partitioning into clusters and thus facilitating faster training. We believe
this is associated with the ``helpful smoothing'' property identified
by~\cite{santurkar2018how} for shallow networks;
ii) the clusters are more stationary, and the stages of cluster formation and
tightening are not as distinct;
iii) the inter-cluster distance and the clusters themselves are larger.


Weight decay's loss scaling mechanism is complementary to other mechanisms
identified in the literature, for instance that it increases the effective
learning rate~\cite{vanlaarhoven2017l2, zhang2019three}. Our results are
consistent with these works in that weight decay reduces the generalization gap
(between training and test error), even in batch-normalized networks where it is
presumed to have no effect. Given that batch norm is not typically used on all
layers, the loss scaling mechanism persists although to a lesser degree in this
case.\looseness=-1

\biblio

\section{Conclusion}
\label{sec:conclusion}

We found that there is no free lunch with batch norm: the accelerated training
properties and occasionally higher clean test accuracy come at the cost of robustness,
both to additive noise and for adversarial perturbations.
We have shown that there is no inherent relationship between the
input dimension and vulnerability. Our results highlight the
importance of identifying the disparate mechanisms of regularization
techniques, especially when concerned about robustness.

\biblio

\section*{Acknowledgements}
The authors wish to acknowledge the financial support of NSERC, CFI, CIFAR and
EPSRC\@. We also acknowledge hardware support from NVIDIA and Compute
Canada. Research at the Perimeter Institute is supported by Industry Canada and
the province of Ontario through the
Ministry of Research \& Innovation.
We thank Thorsteinn Jonsson for helpful discussions; Colin Brennan, Terrance
DeVries and J\"{o}rn-Henrik Jacobsen for technical suggestions;
Justin Gilmer for suggesting the common corruption benchmark;
Maeve Kennedy, Vithursan Thangarasa, Katya
Kudashkina, and Boris Knyazev for comments and proofreading.


\bibliographystyle{abbrv}
\bibliography{icml2019_zotero}

\clearpage
\appendix

\section{Supplement to Empirical Results}
\label{sec:emp-continued}

This section contains supplementary explanations and results to those of
Section~\ref{sec:empirical}.

\subsection{Why the VGG Architecture?}

For SVHN and CIFAR-10 experiments, we selected the VGG family of models as a
simple yet contemporary convolutional
architecture whose development occurred independent of batch norm. This makes it
suitable for a causal intervention, given that we want to study the effect of
batch norm itself, and not batch norm + other architectural innovations +
hyperparameter tuning. State-of-the-art architectures, such as Inception and ResNet,
whose development is more intimately linked with batch norm may be less suitable
for this kind of analysis. The superior standard test accuracy of these models
is somewhat moot given a trade-off between standard test accuracy and robustness,
demonstrated in this work and elsewhere~\cite{tanay2016boundary, galloway2018adversariala,
su2018robustness, tsipras2019robustness}.
Aside from these reasons, and provision of pre-trained variants on ImageNet with
and without batch norm in~\texttt{torchvision.models} for ease of
reproducibility, this choice of architecture is arbitrary.

\subsection{Comparison of PGD to BIM}
\label{sec:pgd-to-bim}

We used the PGD implementation from~\cite{ding2019advertorch} with
settings as below. The pixel range was set to $[-1, 1]$ for SVHN, and
$[-2,2]$ for CIFAR-10 and ImageNet:
\begin{verbatim}
from advertorch.attacks import LinfPGDAttack
adversary = LinfPGDAttack(net, loss_fn=nn.CrossEntropyLoss(reduction="sum"),
    eps=0.03, nb_iter=20, eps_iter=0.003,
    rand_init=False, clip_min=-1.0, clip_max=1.0, targeted=False)
\end{verbatim}

We compared PGD using a step size of $\epsilon / 10$ to our own BIM implemenation
with a step size of $\epsilon / 20$, for the same number (20) of iterations.
This reduces test accuracy for $\epsilon_\infty=0.03$ perturbations
from $31.3 \pm 0.2\%$ for BIM to $27.1 \pm 0.3\%$ for PGD for the unnormalized
VGG8 network, and from $15 \pm 1\%$ to $10 \pm 1\%$ for the batch-normalized
network. The difference due to batch norm is identical in both cases:
$17 \pm 1\%$. Results were also consistent between PGD and BIM for ImageNet.
We also tried increasing the number of PGD iterations for deeper
networks. For VGG16 on CIFAR-10, using 40 iterations of PGD with a step size of
$\epsilon_\infty / 20$, instead of
20 iterations with $\epsilon_\infty / 10$, reduced accuracy from
$28.9 \pm 0.2\%$ to $28.5 \pm 0.3\%$, a difference of only $0.4 \pm 0.5\%$.

\subsection{Additional SVHN and CIFAR-10 Results for Deeper Models}
\label{sec:appendix-deeper}

Our first attempt to train VGG models on SVHN with more than 8 layers failed,
therefore for a fair comparison we report the
robustness of the deeper models that were only trainable by using batch norm in
Table~\ref{tab:svhn-depth-bn}. None of these models obtained much better
robustness in terms of PGD-$\ell_2$, although they did better for PGD-$\ell_\infty$.

\begin{table}[h]
\centering
\caption{VGG variants on SVHN with batch norm.}
\begin{tabular}{ccccc}
\toprule
\multicolumn{1}{c}{} & \multicolumn{4}{c}{Test Accuracy ($\%$)} \\ \midrule
L & Clean & Noise & PGD-$\ell_\infty$ & PGD-$\ell_2$ \\ \midrule
11 & $95.31 \pm 0.03$ & $80.5 \pm 1$ & $20.2 \pm 0.2$ & $6.1 \pm 0.2$ \\ \midrule
13 & $95.88 \pm 0.05$ & $77.2 \pm 7$ & $21.7 \pm 0.5$ & $5.4 \pm 0.2$ \\ \midrule
16 & $94.59 \pm 0.05$ & $78.1 \pm 4$ & $19.2 \pm 0.3$ & $3.0 \pm 0.2$ \\ \midrule
19 & $95.1 \pm 0.3$ & $78 \pm 1$ & $24.2 \pm 0.6$ & $4.1 \pm 0.4$ \\ \bottomrule
\end{tabular}
\label{tab:svhn-depth-bn}
\end{table}

Fixup initialization was recently proposed to reduce
the use of normalization layers in deep residual networks~\cite{zhang2019residual}.
As a natural test we compare a WideResNet (28 layers, width factor 10) with Fixup
versus the default architecture with batch norm. Note that the Fixup variant
still contains one batch norm layer before the classification layer, but the
number of batch norm layers is still greatly reduced.\footnote{We used the
implementation from~\url{https://github.com/valilenk/fixup}, but stopped
training at 150 epochs for consistency with the VGG8 experiment. Both
models had already fit the training set by this point.}

\begin{table}[h]
\centering
\caption{Accuracies of WideResNet--28--10 on CIFAR-10 and CIFAR-10.1 (\texttt{v6}).}
\begin{tabular}{ccccccc}
\toprule
\multicolumn{1}{c}{} & \multicolumn{4}{c}{CIFAR-10} &
\multicolumn{2}{c}{CIFAR-10.1} \\ \midrule
Model & Clean & Noise & PGD-$\ell_\infty$ & PGD-$\ell_2$ & Clean & Noise \\ \midrule
Fixup & $94.6 \pm 0.1$ & $69.1 \pm 1.1$ & $20.3 \pm 0.3$ & $9.4 \pm 0.2$
& $87.5 \pm 0.3$ & $67.8 \pm 0.9$\\ \midrule
BN & $95.9 \pm 0.1$ & $57.6 \pm 1.5$ & $14.9 \pm 0.6$ & $8.3 \pm 0.3$
& $89.6 \pm 0.2$ & $58.3 \pm 1.2$ \\ \bottomrule
\end{tabular}
\label{tab:cifar10-wrn}
\end{table}

We train WideResNets (WRN) with five unique seeds and show their test accuracies
in Table~\ref{tab:cifar10-wrn}.
Consistent with~\cite{recht2018cifar10a}, higher clean test accuracy on CIFAR-10,
i.e.~obtained by the WRN compared to VGG, translated to higher clean accuracy on
CIFAR-10.1. However, these gains were wiped out by moderate Gaussian noise.
VGG8 dramatically outperforms both WideResNet variants
subject to noise, achieving $78.9 \pm 0.6$ vs.~$69.1 \pm 1.1$.
Unlike for VGG8, the WRN showed little generalization gap between noisy
CIFAR-10 and 10.1 variants: $69.1 \pm 1.1$ is reasonably compatible with
$67.8 \pm 0.9$, and $57.6 \pm 1.5$ with $58.3 \pm 1.2$.
The Fixup variant improves accuracy by $11.6 \pm 1.9\%$ for noisy CIFAR-10,
$9.5 \pm 1.5\%$ for noisy CIFAR-10.1, $5.4 \pm 0.6\%$ for PGD-$\ell_\infty$, and
$1.1 \pm 0.4\%$ for PGD-$\ell_2$.

We believe our work serves as a compelling motivation for
Fixup and other techniques that aim to reduce usage of batch normalization.
The role of skip-connections should be isolated in future work since absolute
values were consistently lower for residual networks.

\subsection{ImageNet Black-box Transferability Analysis}
\label{sec:imagenet-black-box}

\begin{table}[h]
\centering
\caption{ImageNet validation accuracy for adversarial examples transfered
between VGG variants of various depths, indicated by number, with and without
batch norm (``\cmark'', ``\xmark''). All adversarial
examples were crafted with BIM-$\ell_\infty$ using 10 steps and a step size of
5e-3, which is higher than for the white-box analysis to improve transferability.
The BIM objective was simply misclassification, i.e., it was not a targeted attack.
For efficiency reasons, we select 2048 samples from the validation set.
Values along the diagonal in first two columns for Source = Target indicate
white-box accuracy.}
\begin{tabular}{ccccccccccc}
\toprule
\multicolumn{2}{c}{\multirow{2}{*}{}} & \multicolumn{8}{c}{Target} \\ \midrule
& \multicolumn{2}{c}{} & \multicolumn{2}{c}{11} & \multicolumn{2}{c}{13} & \multicolumn{2}{c}{16} & \multicolumn{2}{c}{19} \\ \midrule
Acc.~Type & \multicolumn{2}{c}{Source} & \xmark & \cmark & \xmark & \cmark & \xmark & \cmark & \xmark & \cmark \\ \midrule
\multirow{2}{*}{Top 1} & \multirow{2}{*}{11} & \xmark & 1.2 & 42.4 & 37.8 & 42.9 & 43.8 & 49.6 & 47.9 & 53.8 \\ \cmidrule{3-11}
 & & \cmark & 58.8 & 0.3 & 58.2 & 45.0 & 61.6 & 54.1 & 64.4 & 58.7 \\ \bottomrule
\toprule
\multirow{2}{*}{Top 5} & \multirow{2}{*}{11} & \xmark & 11.9 & 80.4 & 75.9 & 80.9 & 80.3 & 83.3 & 81.6 & 85.1 \\ \cmidrule{3-11}
  & & \cmark & 87.9 & 6.8 & 86.7 & 83.7 & 89.0 & 85.7 & 90.4 & 88.1 \\ \bottomrule
\end{tabular}
\label{tab:imagenet-black-box}
\end{table}

The discrepancy between the results in additive noise and for white-box BIM
perturbations for ImageNet in Section 3 raises a natural
question: Is~\emph{gradient masking} a factor influencing the success of the
white-box results on ImageNet? No, consistent with the white-box results,
when the target is unnormalized but the source is, top 1 accuracy is
$10.5\% - 16.4\%$ higher, while top 5 accuracy is $5.3\% - 7.5\%$ higher, than
vice versa. This can be observed in Table~\ref{tab:imagenet-black-box} by
comparing the diagonals from lower left to upper right.
When targeting an unnormalized model, we reduce top 1 accuracy by
$16.5\% - 20.4\%$ using a source that is also unnormalized, compared to a
difference of only $2.1\% - 4.9\%$ by matching batch normalized networks.
This suggests that the features used by unnormalized networks are more stable
than those of batch normalized networks.

Unfortunately, the pre-trained ImageNet models provided by the PyTorch developers
do not include hyperparameter settings or other training details. However, we
believe that this speaks to the generality of the results, i.e., that they are
not sensitive to hyperparameters.

\subsection{Batch Norm Limits Maximum Trainable Depth and Robustness}
\label{sec:batch-norm-batch-size}

\begin{figure}
\centering
\subfigure[]{\includegraphics[width=0.45\columnwidth]{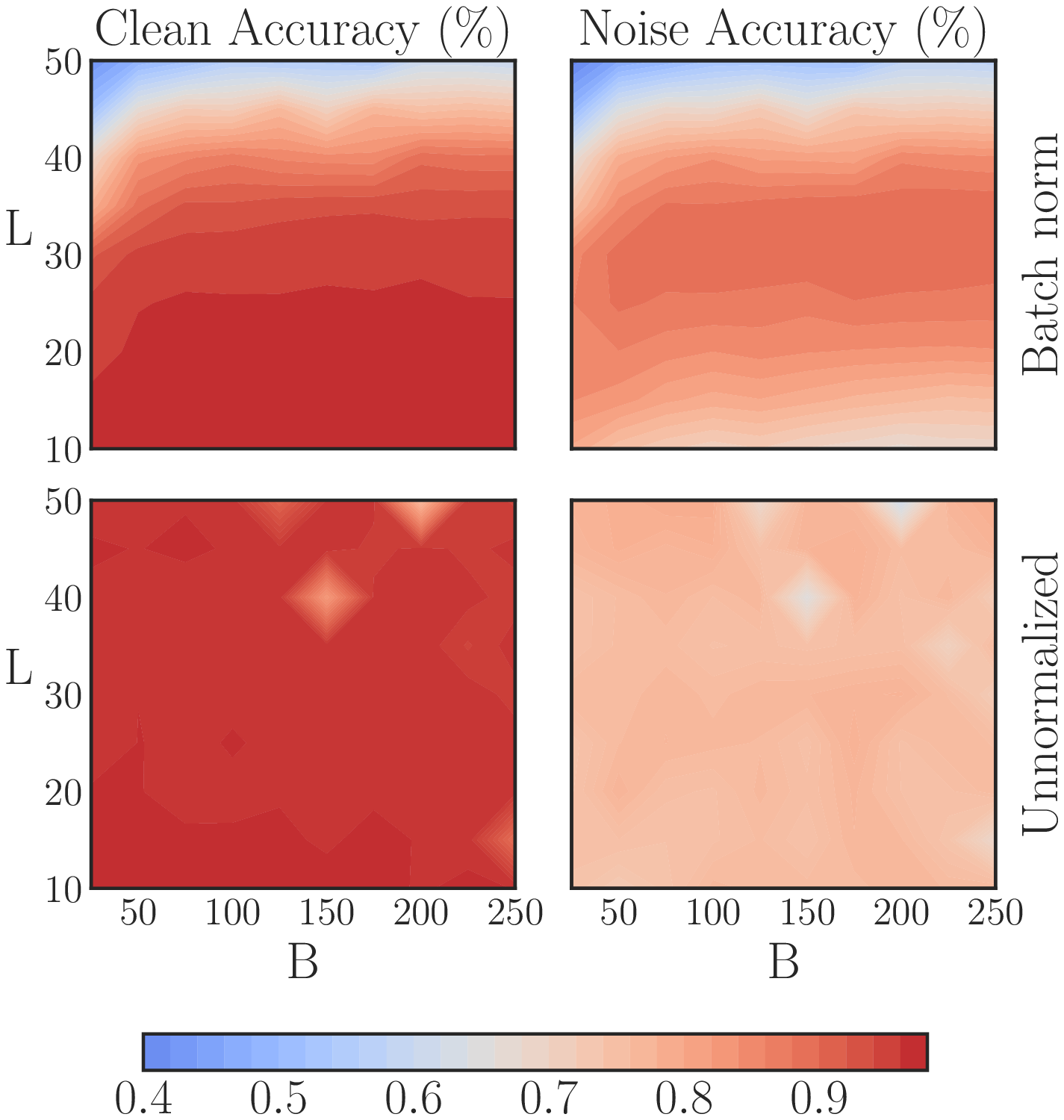}
\label{fig:ep10}}
\subfigure[]{\includegraphics[width=0.45\columnwidth]{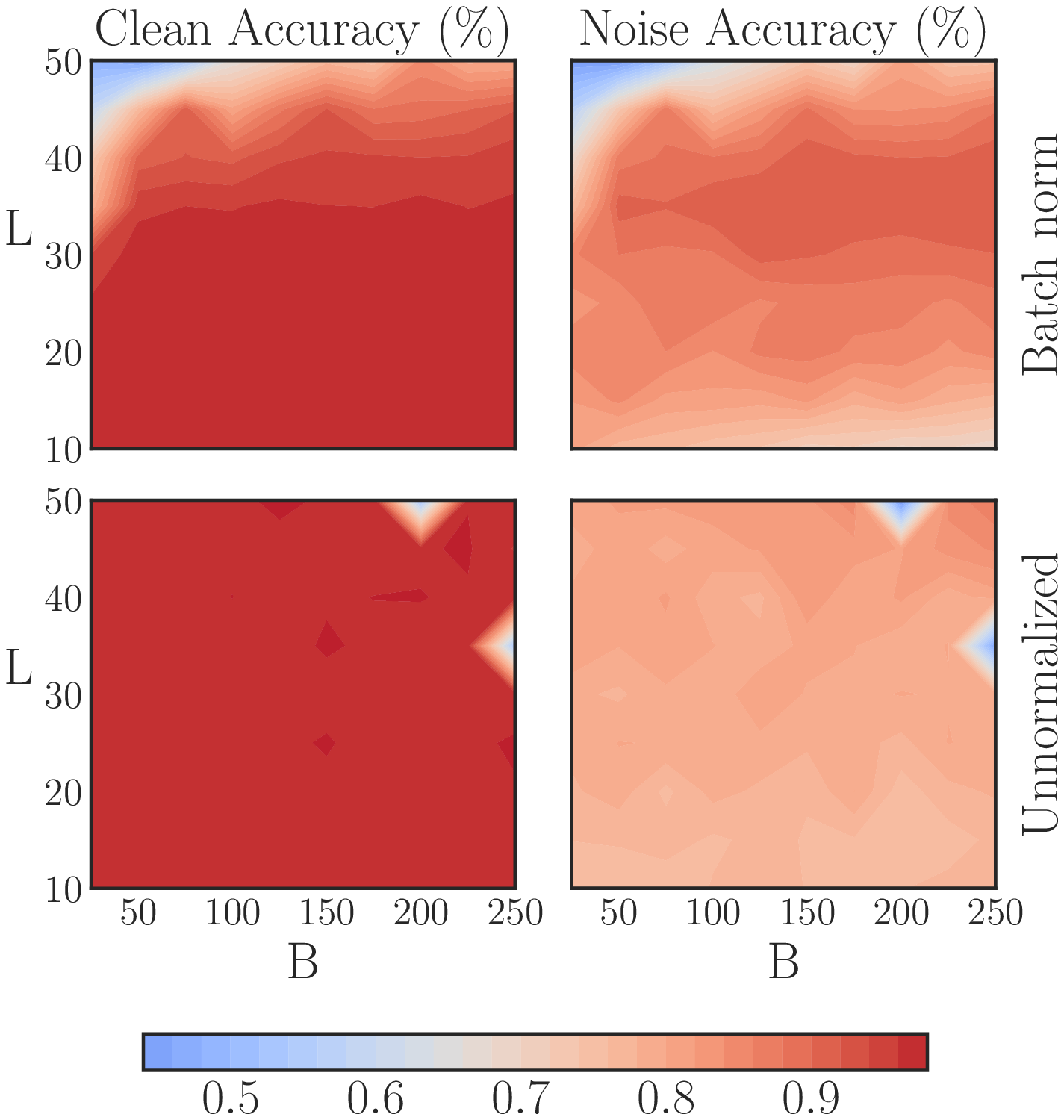}
\label{fig:ep40}}
\caption{We repeat the experiment of~\cite{yang2019mean} by training
fully-connected models of depth $L$ and constant width ($N_l$=384) with ReLU
units by SGD, and learning rate $\eta = 10^{-5} B$ for batch size $B$ on MNIST\@.
We train for 10 and 40 epochs in~\subref{fig:ep10} and~\subref{fig:ep40}
respectively.
The batch norm parameters $\gamma$ and $\beta$ were left as default, momentum
disabled, and $c$ = 1e-3.
Each
coordinate is first averaged over three seeds. Diamond-shaped artefacts
for unnormalized case indicate one of three seeds failed to train -- note that 
we show an equivalent version of~\subref{fig:ep10} with these outliers removed
and additional batch
sizes from 5--20 in Figure 2. Best viewed in colour.}
\label{fig:mft-batch-appendix}
\end{figure}

In Figure~\ref{fig:mft-batch-appendix} we show that batch norm not only limits the
maximum trainable depth, but robustness decreases with the batch size for depths
that maintain test accuracy, at around 25 or fewer layers
(in Figure~\ref{fig:ep10}). Both clean accuracy
and robustness showed little to no relationship with depth nor batch size in
unnormalized networks. A few outliers are observed for unnormalized networks at
large depths and batch size, which could be due to the reduced number of
parameter update steps that result from a higher batch size and fixed number of
epochs~\cite{hoffer2017train}.

Note that in Figure~\ref{fig:ep10} the bottom row---without batch norm---appears
lighter than the equivalent plot above, with batch norm, indicating that
unnormalized networks obtain less absolute peak accuracy than the batch-normalized
network. Given that the unnormalized networks take longer to
converge, we prolong training for 40 epochs total.
When they do converge, we see more configurations that achieve higher
clean test accuracy than batch-normalized networks in Figure~\ref{fig:ep40}.
Furthermore, good robustness can be experienced simultaneously with good
clean test accuracy in unnormalized networks, whereas the regimes of
good clean accuracy and robustness are still mostly non-overlapping in
Figure~\ref{fig:ep40}.

\biblio

\section{Weight Decay and Input Dimension}
\label{sec:appendix-input-dimension}

Consider a logistic classification model represented by a neural network consisting of a single unit,
parameterized by weights $w\in \mathbb{R}^d$ and bias $b\in \mathbb{R}$,
with input denoted by $x \in \mathbb{R}^d$ and true labels $y \in \{\pm1\}$.
Predictions are defined by $s = w^\top x + b$, and the model is optimized through empirical risk minimization,
i.e., by applying stochastic gradient descent (SGD) to the loss function~\eqref{eq:objective1},
where $\zeta(z) = \log(1 + e^{-z})$:
\begin{equation}
	\mathbb{E}_{x, y \sim p_\text{data}}\; \zeta(y (w^\top x + b)).
\label{eq:objective1}
\end{equation}

We note that $\;w^\top x + b\;$ is a \emph{scaled}, signed distance between $x$
and the classification boundary defined by our model. If we define $d(x)$ as the
signed Euclidean distance between $x$ and the boundary,
then we have: $w^\top x + b = \|w\|_2\,d(x)$. Hence,
minimizing~\eqref{eq:objective1} is equivalent to minimizing
\begin{equation}
\mathbb{E}_{x, y \sim p_\text{data}}\; \zeta(\|w\|_2 \times y\,d(x)).
\label{eq:objective2}
\end{equation}
We define the \emph{scaled loss} as
\begin{equation}
\zeta_{\|w\|_2}(z) := \zeta(\|w\|_2 \times z)
\label{eq:scaled-softplus}
\end{equation}
and note that adding a $\ell_2$ regularization term in~\eqref{eq:objective2},
resulting in~\eqref{eq:objective3}, can be understood as a way of controlling the
scaling of the loss function:
\begin{equation}
\mathbb{E}_{x, y \sim p_\text{data}}\; \zeta_{\|w\|_2}(y\,d(x)) + \lambda\|w\|_2 
\label{eq:objective3}
\end{equation}


\newcommand{\cellwidth}{.8in}

\begin{figure*}[h]
\begin{center}
\centering{\subfigure[$w^\top x$]{
\includegraphics[width=\cellwidth]{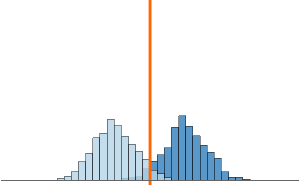}
\label{fig:wd_vs_at_cell1}}} 
\centering{\subfigure[$y (w^\top x)$]{
\includegraphics[width=\cellwidth]{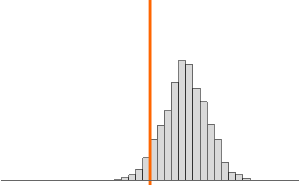}
\label{fig:wd_vs_at_cell2}}} 
\centering{\subfigure[$\zeta(y (w^\top x))$]{
\includegraphics[width=\cellwidth]{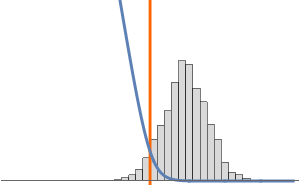}
\label{fig:wd_vs_at_cell3}}} 
\centering{\subfigure[$\zeta_{5}(y\,d(x))$]{
\includegraphics[width=\cellwidth]{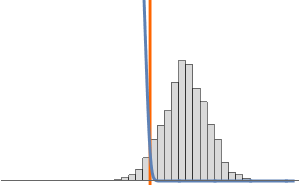}
\label{fig:wd_vs_at_cell4}}} 
\centering{\subfigure[$\zeta_{0.5}(y\,d(x))$]{
\includegraphics[width=\cellwidth]{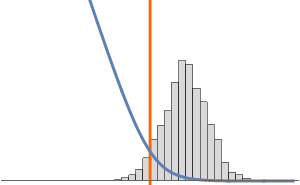}
\label{fig:wd_vs_at_cell5}}} 
\centering{\subfigure[$\zeta_{0.05}(y\,d(x))$]{
\includegraphics[width=\cellwidth]{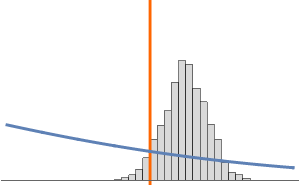}
\label{fig:wd_vs_at_cell6}}} 
\end{center}
\caption{\subref{fig:wd_vs_at_cell1} For a given weight vector $w$ and bias $b$,
the values of $\,w^\top x + b\,$ over the training set typically follow a
bimodal distribution (corresponding to the two classes) centered on the
classification boundary. \subref{fig:wd_vs_at_cell2}  Multiplying by the label
$y$ allows us to distinguish the correctly classified data in the positive
region from misclassified data in the negative
region.~\subref{fig:wd_vs_at_cell3} We can then attribute a penalty to each
training point by applying the loss to $y(w^\top x + b)$.
\subref{fig:wd_vs_at_cell4} For a small regularization parameter
(large $\|w\|_2$), the misclassified data is penalized linearly while the
correctly classified data is not penalized.~\subref{fig:wd_vs_at_cell5} A
medium regularization parameter (medium $\|w\|_2$) corresponds to smoothly
blending the margin.~\subref{fig:wd_vs_at_cell6} For a large regularization
parameter (small $\|w\|_2$), all data points are penalized almost linearly.}
\label{fig:wd_vs_at1}
\end{figure*}

In Figures~\ref{fig:wd_vs_at_cell1}-\ref{fig:wd_vs_at_cell3}, we develop intuition
for the different quantities contained in~\eqref{eq:objective1} with respect to
a typical binary classification problem, while
Figures~\ref{fig:wd_vs_at_cell4}-\ref{fig:wd_vs_at_cell6} depict the effect
of the regularization parameter $\lambda$ on the scaling of the loss function.

To test this theory empirically we study a model with a single linear layer
(number of units equals input dimension) and cross-entropy loss function
on variants of MNIST of increasing input dimension,
to approximate the toy model described in the ``core idea'' from~\cite{simon-gabriel2018adversarial}
as closely as possible, but with a model capable of learning. Clearly, this model is too
simple to obtain competitive test accuracy, but this is a helpful first step that
will be subsequently extended to ReLU networks.
The model was trained by SGD for 50 epochs with a constant learning rate of 1e-2
and a mini-batch size of 128.
In Table~\ref{tab:linear-model-wd} we show that increasing the input dimension by
resizing MNIST from $28 \times 28$ to various resolutions with~\texttt{PIL.Image.NEAREST}
interpolation increases adversarial vulnerability in terms of accuracy and loss.
Furthermore, the ``adversarial damage'', defined as the average increase of the loss after attack,
which is predicted to grow like $\sqrt{d}$ by Theorem 4 of~\cite{simon-gabriel2018adversarial},
falls in between that obtained empirically for $\epsilon = 0.05$ and $\epsilon = 0.1$ for all image widths
except for 112, which experiences slightly more damage than anticipated.

\cite{simon-gabriel2018adversarial} note that independence between vulnerability
and the input dimension can be recovered through adversarial-example augmented
training by projected gradient descent (PGD), with a small trade-off in terms of
standard test accuracy.
We find that the same can be achieved through a much simpler approach: $\ell_2$ weight
decay, with parameter $\lambda$ chosen dependent on $d$ to correct for the loss scaling.
This way we recover input dimension invariant vulnerability with little
degradation of test accuracy, e.g., see the result for $\sqrt{d} = 112$ and $\epsilon = 0.1$ in
Table~\ref{tab:linear-model-wd}: the accuracy ratio is $1.00 \pm 0.03$ with weight decay regularization, 
compared to $0.10 \pm 0.09$ without.

Compared to PGD training, weight decay regularization i) does not have an 
arbitrary $\epsilon$
hyperparameter that ignores inter-sample distances,
ii) does not prolong training by a multiplicative factor given by the 
number of steps in the inner loop, and 3) is less attack-specific. Thus, we do 
not use adversarially augmented
training because we wish to convey a notion of robustness to unseen
attacks and common corruptions. Furthermore, enforcing robustness to
$\epsilon$-perturbations may increase vulnerability to~\emph{invariance-based}
examples, where semantic changes are made to the input, thus changing the Oracle
label, but not the classifier's prediction~\cite{jacobsen2019exploiting}.
Our models trained with
weight decay obtained $12\%$ higher accuracy (86\% vs.~74\% correct) compared to
batch norm on a small sample of 100 $\ell_\infty$ invariance-based MNIST
examples.\footnote{Invariance
based adversarial examples downloaded
from~\url{https://github.com/ftramer/Excessive-Invariance}.}
We make primary use of traditional $\ell_p$ perturbations as
they are well studied in the literature and straightforward to compute, but
solely defending against these is not the end goal.

A more detailed
comparison between adversarial training and weight decay can be found
in~\cite{galloway2018adversariala}. The scaling of the loss function mechanism
of weight decay is complementary to other mechanisms identified in the literature
recently, for instance that it also increases the effective learning
rate~\cite{vanlaarhoven2017l2, zhang2019three}. Our results are consistent with
these works in that weight decay reduces the generalization gap, even in batch-normalized
networks where it is presumed to have no effect. Given that batch norm is not
typically used on the last layer, the loss scaling mechanism persists in this
setting, albeit to a lesser degree.

\begin{table}[]
\centering
\caption{Mitigating the effect of the input dimension on adversarial
vulnerability by correcting the margin enforced by the loss function.
Regularization constant $\lambda$ is for $\ell_2$ weight decay.
Consistent with~\cite{simon-gabriel2018adversarial}, we use
$\epsilon$-FGSM perturbations, the optimal $\ell_\infty$ attack for a
linear model. Values in rows with
$\sqrt{d} > 28$ are ratios of entry (accuracy or loss) wrt the $\sqrt{d} = 28$
baseline. ``Pred.''~is the predicted increase of the loss $\mathcal{L}$
due to a small $\epsilon$-perturbation using Thm.~4 of~\cite{simon-gabriel2018adversarial}.}

\begin{tabular}{cccccccc}
\toprule
\multicolumn{2}{c}{Model} & \multicolumn{2}{c}{(Relative) Test Accuracy} & \multicolumn{3}{c}{(Relative) Loss} \\
\midrule
$\sqrt{d}$ & $\lambda$ 	& Clean 				& $\epsilon = 0.1$ 		& Clean 				& $\epsilon = 0.1$ & Pred.\\ \midrule
28 & -- 				& $92.4 \pm 0.1\%$ 		& $53.9 \pm 0.3\%$ 		& $0.268 \pm 0.001$ 	& $1.410 \pm 0.004$ & -\\ \midrule \midrule
56 & -- 				& $1.001 \pm 0.001$ 	& $0.33 \pm 0.03$ 		& $1.011 \pm 0.007$ 	& $2.449 \pm 0.009$ & 2 \\ \midrule
56 & 0.01 				& $0.999 \pm 0.002$ 	& $0.98 \pm 0.01$ 		& $1.010 \pm 0.007$ 	& $1.01 \pm 0.01$ & - \\ \midrule
84 & -- 				& $0.998 \pm 0.002$ 	& $0.10 \pm 0.09$ 		& $1.06 \pm 0.01$ 		& $4.15 \pm 0.02$ & 3 \\ \midrule
84 & 0.0225 			& $0.996 \pm 0.003$ 	& $0.96 \pm 0.04$ 		& $1.05 \pm 0.02$ 		& $1.06 \pm 0.03$ & - \\ \midrule
112 & -- 				& $0.992 \pm 0.004$ 	& $0.1 \pm 0.2$ 		& $1.18 \pm 0.03$ 		& $5.96 \pm 0.02$ & 4 \\ \midrule
112 & 0.05 			& $0.987 \pm 0.004$ 	& $1.00 \pm 0.03$ 		& $1.14 \pm 0.04$ 		& $1.04 \pm 0.03$ & - \\ \bottomrule
\end{tabular}
\label{tab:linear-model-wd}
\end{table}
%

\begin{table}[]
\centering
\caption{Two-hidden-layer ReLU MLP (see main text for architecture), with and
without batch norm (BN), trained
for 50 epochs and repeated over five random seeds.
Values in rows with $\sqrt{d} > 28$ are ratios wrt the $\sqrt{d} = 28$ baseline
(accuracy or loss).
There is a considerable increase of the loss, or similarly, a degradation of
robustness in terms of accuracy, due to batch norm. The discrepancy for
BIM-$\ell_\infty$ with $\epsilon = 0.1$ for $\sqrt{d} = 84$ with
batch norm represents a $61 \pm 1\%$ degradation in absolute accuracy compared
to the baseline.}
\begin{tabular}{cccccccc}
\toprule
\multicolumn{2}{c}{Model} & \multicolumn{2}{c}{(Relative) Test Accuracy} & \multicolumn{2}{c}{(Relative) Loss} \\ \midrule
$\sqrt{d}$ & BN & Clean 				& $\epsilon = 0.1$ 	& Clean 				& $\epsilon = 0.1$ \\ \midrule
28 & \xmark 	& $97.95 \pm 0.08\%$ 	& $66.7 \pm 0.9 \%$ 	& $0.0669 \pm 0.0008$ 	& $1.06\pm 0.02$\\ \midrule \midrule
28 & \cmark 	& $0.9992 \pm 0.0012$ 	& $0.34 \pm 0.03$ 	& $1.06 \pm 0.04$ 		& $3.18\pm 0.03$\\ \midrule 
56 & \xmark 	& $1.0025 \pm 0.0009$ 	& $0.80 \pm 0.02$ 	& $0.87 \pm 0.02$ 		& $1.68 \pm 0.03$\\ \midrule
56 & \cmark 	& $1.0027 \pm 0.0008$ 	& $0.13 \pm 0.09$ 	& $0.91 \pm 0.03$ 		& $5.83 \pm 0.03$\\ \midrule 
84 & \xmark 	& $1.0033 \pm 0.0009$ 	& $0.71 \pm 0.02$ 	& $0.86 \pm 0.02$ 		& $2.15 \pm 0.03$\\ \midrule
84 & \cmark 	& $1.0033 \pm 0.0010$ 	& $0.09 \pm 0.08$ 	& $0.88 \pm 0.02$ 		& $7.34 \pm 0.02$\\ \bottomrule
\end{tabular}
\label{tab:three-layer-bn-full-ratio}
\end{table}

\biblio

\section{Adversarial Spheres}
\label{sec:spheres}

\begin{figure}
\centering
\subfigure[]{
\includegraphics[width=.48\columnwidth]{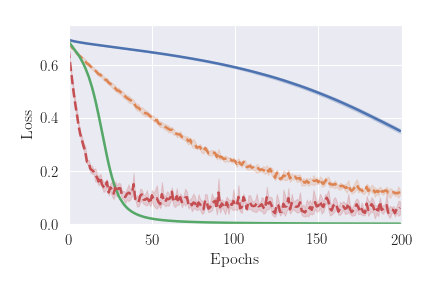}
\label{fig:train-lss}}
\subfigure[]{
\includegraphics[width=.48\columnwidth]{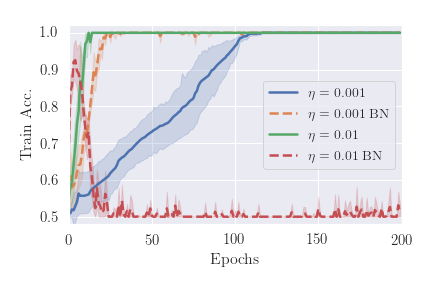}
\label{fig:train-acc}}
\caption{We train the same two-hidden-layer fully connected network of
width 1000 units using ReLU activations and a mini-batch size of 50 on a 2D
variant of the ``Adversarial Spheres'' binary classification
problem~\cite{gilmer2018adversarial}. Dashed lines denote the model with batch
norm. The batch-normalized model fails to train for a learning rate of $\eta=0.01$,
which otherwise converges quickly for the unnormalized equivalent. We repeat the
experiment over five random seeds, shaded regions indicate a $95\%$ confidence
interval.}
\label{fig:train-stats-2d}
\end{figure}

The ``Adversarial Spheres'' dataset contains points sampled uniformly from the surfaces of
two concentric $n$-dimensional spheres with radii $R=1$ and $R=1.3$ respectively, and
the classification task is to attribute a given point to the inner or outer sphere. We consider the
case $n=2$, that is, datapoints from two concentric circles.
This simple problem poses a challenge to the conventional wisdom regarding batch norm:
not only does batch norm harm robustness, it makes training less stable.
In Figure~\ref{fig:train-stats-2d} we show that, using the same architecture as
in~\cite{gilmer2018adversarial}, the batch-normalized network is highly
sensitive to the learning rate $\eta$.
We use SGD instead of Adam to avoid introducing unnecessary complexity,
and especially since SGD has been shown to converge to the maximum-margin
solution for linearly separable data~\cite{soudry2018implicita}. We use a finite
dataset of 500 samples from $\mathcal{N}(0, I)$ projected onto the circles.
The unormalized network achieves zero training error for $\eta$ up to 0.1 (not
shown), whereas the batch-normalized network is already untrainable at
$\eta = 0.01$.
To evaluate robustness, we sample 10,000 test points from the
same distribution for each class (20k total), and apply noise drawn from
$\mathcal{N}(0, 0.005 \times I)$. We evaluate only the models that could be
trained to $100\%$ training accuracy with the smaller learning rate of
$\eta = 0.001$. The model with batch norm classifies $94.83\%$ of these
points correctly, while the unnormalized net obtains $96.06\%$.

\begin{figure}
\centering
\subfigure[]{
\includegraphics[width=.45\columnwidth]{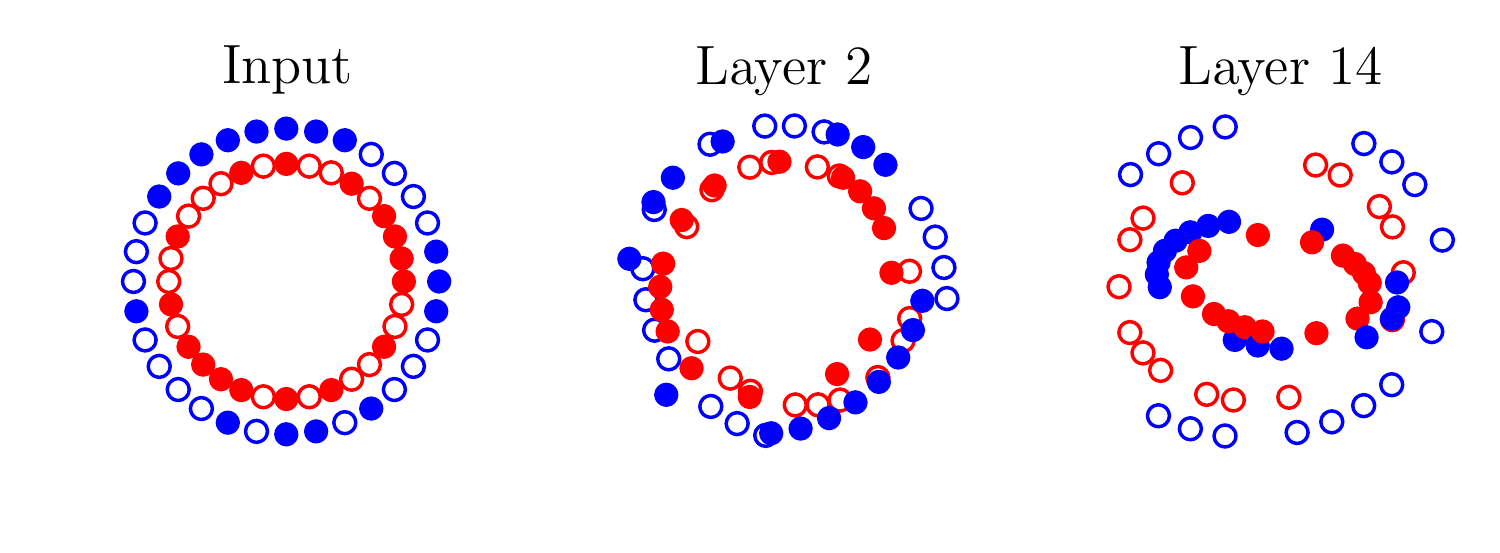}
\label{fig:sp-bn}}
\subfigure[]{
\includegraphics[width=.45\columnwidth]{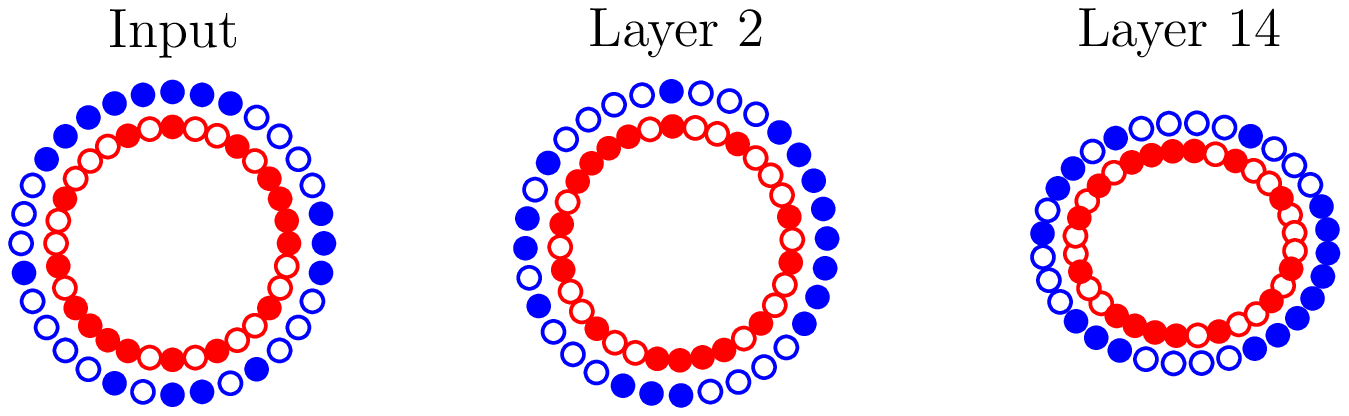}
\label{fig:sp-no-bn}}
\caption{Two mini-batches from the ``Adversarial Spheres'' dataset (2D variant), and their representations in a deep
linear network at initialization time~\subref{fig:sp-bn} with batch norm and~\subref{fig:sp-no-bn} without batch norm.
Mini-batch membership is indicated by marker fill and class membership
by colour. Each layer is projected to its two principal components.
In~\subref{fig:sp-no-bn} we scale both components by a factor of 100, as the
dynamic range decreases with depth under default initialization.
We observe in~\subref{fig:sp-bn} that some samples are already overlapping at
Layer 2, and classes are mixed at Layer 14.}
\label{fig:app-adv-circle}
\end{figure}

\biblio

\section{Author Contributions}
\label{sec:contributions}

In the spirit of~\cite{sculley2018winnera}, we provide a summary of
each author's contributions.
\begin{itemize}
  \item First author formulated the hypothesis, conducted
  the experiments, and wrote the initial draft.
  \item Second author prepared detailed technical notes on the main references,
  met frequently with the first author to advance the work, and critically revised
  the manuscript.
  \item Third author originally conceived the key theoretical concept of
  Appendix~\ref{sec:appendix-input-dimension} as well as some of the figures,
  and provided important technical suggestions and feedback.
  \item Fourth author met with the first author to discuss the work and helped
  revise the manuscript.
  \item Senior author critically revised several iterations of the manuscript,
  helped improve the presentation, recommended additional experiments, and
  sought outside feedback.
\end{itemize}

\biblio

\end{document}


\section{Common Corruption Robustness}
\label{sec:common-corruptions}

We evaluated robustness on the common corruptions and perturbations
benchmarks~\cite{hendrycks2019benchmarking}.
Common corruptions are 19 types of real-world effects that can be grouped
into four categories: ``noise'', ``blur'', ``weather'',
and ``digital''. Each corruption has five ``severity'' or
intensity levels. These are applied to the test sets of CIFAR-10 and
ImageNet, denoted CIFAR-10-C and ImageNet-C respectively.
When reporting the mean corruption error (mCE), we average over intensity levels
for each corruption, then over all corruptions.
We outline the results for two VGG variants and a WideResNet on
CIFAR-10-C, trained from scratch independently over three and five random seeds,
respectively. The most important results are also summarized in
Table~\ref{tab:cifar10-c}.

\begin{table}[]
\centering
\caption{Robustness of three modern convolutional neural network
architectures with and without batch norm on the
\texttt{CIFAR-10-C} common corruptions benchmark~\cite{hendrycks2019benchmarking}.
We use ``F'' to denote Fixup~\cite{zhang2019residual}, as this variant still
had one batch-norm layer. Values were averaged over five intensity levels for
each corruption. We report the top five of 19 corruptions by
magnitude of the accuracy gap due to batch norm; see the text for more detail on
the corruptions omitted here.}
\begin{tabular}{cccccccc}
\toprule
\multicolumn{2}{c}{Model} & \multicolumn{6}{c}{Test Accuracy ($\%$)} \\ \midrule
Variant & BN & Clean & Gaussian & Impulse & Shot & Speckle & Contrast \\ \midrule
\multirow{2}{*}{VGG8} & \xmark & $87.9 \pm 0.1$ & $\mathbf{65.6 \pm 1.2}$
 & $\mathbf{58.8 \pm 0.8}$ & $\mathbf{71.0 \pm 1.2}$ & $\mathbf{70.8 \pm 1.2}$
 & $\mathbf{59.3 \pm 0.8}$\\ \cline{2-8}
 & \cmark & $88.7\pm 0.1$ & $56.4 \pm 1.5$ & $51.2 \pm 0.1$ & $65.4 \pm 1.1$
 & $66.3 \pm 1.1$ & $54.9 \pm 1.0$ \\ \midrule 
\multirow{2}{*}{VGG13} & \xmark & $91.74 \pm 0.02$ & $\mathbf{64.5\pm 0.8}$
 & $\mathbf{63.3 \pm 0.3}$ & $\mathbf{70.9 \pm 0.4}$ & $\mathbf{71.5 \pm 0.5}$
 & $65.3 \pm 0.6$\\ \cline{2-8}
 & \cmark & $93.0 \pm 0.1$ & $43.6 \pm 1.2$ & $49.7 \pm 0.5$ & $56.8 \pm 0.9$
 & $60.4 \pm 0.7$ & $\mathbf{67.7 \pm 0.5}$ \\ \midrule 
\multirow{2}{*}{WRN28} & F & $94.6 \pm 0.1$ & $\mathbf{63.3 \pm 0.9}$
 & $\mathbf{66.7 \pm 0.9}$ & $\mathbf{71.7 \pm 0.7}$ & $\mathbf{73.5 \pm 0.6}$
 & $81.2 \pm 0.7$ \\ \cline{2-8}
 & \cmark & $95.9 \pm 0.1$ & $51.2 \pm 2.7$ & $56.0 \pm 2.7$ & $63.0 \pm 2.5$
 & $66.6 \pm 2.5$ & $\mathbf{86.0 \pm 0.9}$ \\ \bottomrule

\end{tabular}
\label{tab:cifar10-c}
\end{table}

For VGG8 batch norm increased the error rate
for all noise variants, at every intensity level.
The mean generalization gaps for noise were:
Gaussian---$9.2 \pm 1.9\%$, Impulse---$7.5 \pm 0.8\%$,
Shot---$5.6 \pm 1.6\%$, and Speckle---$4.5 \pm 1.6\%$.
The next most impactful corruptions were:
Contrast---$4.4 \pm 1.3\%$, Spatter---$2.4 \pm 0.7\%$,
JPEG---$2.0 \pm 0.4\%$, and Pixelate---$1.3 \pm 0.5\%$. Results for the remaining
corruptions were a coin toss as to whether batch norm improved or degraded
robustness, as the random error was in the same ballpark as the difference being
measured. These were:
Weather---Brightness, Frost, Snow, and Saturate; Blur---Defocus, Gaussian,
Glass, Zoom and Motion; and Elastic transformation. Averaging over all
corruptions we get an mCE gap of $1.9 \pm 0.9\%$ due to batch norm, or a loss of
accuracy from $72.9 \pm 0.7\%$ to $71.0 \pm 0.6\%$.

VGG13 results were mostly consistent with VGG8: batch norm increased the error rate
for all noise variants, at every intensity level. Particularly notable, the
generalization gap enlarged to $26-28\%$ for Gaussian noise at severity
levels 3, 4, and 5; and $17\%+$ for Impulse noise at levels 4 and 5.
Averaging over all levels, we have gaps for noise variants of:
Gaussian---$20.9 \pm 1.4\%$, Impulse---$13.6 \pm 0.6\%$,
Shot---$14.1 \pm 1.0\%$, and Speckle---$11.1 \pm 0.8\%$. Robustness to the other
corruptions seemed to benefit from the slightly higher clean test accuracy
of $1.3 \pm 0.1\%$ due to batch norm for VGG13. The remaining
generalization gaps varied from (negative) $0.2 \pm 1.3\%$ for Zoom blur, to
$2.9 \pm 0.6\%$ for Pixelate.
Overall mCE was reduced by $2.0 \pm 0.3\%$ for the unnormalized network.

For a WideResNet 28--10 (WRN) using ``Fixup''
initialization~\cite{zhang2019residual} to reduce the use of batch norm, the
mCE was similarly reduced by $1.6 \pm 0.4\%$.
Unpacking each category, the mean generalization gaps for noise were:
Gaussian---$12.1 \pm 2.8\%$, Impulse---$10.7 \pm 2.9\%$,
Shot---$8.7 \pm 2.6\%$, and Speckle---$6.9 \pm 2.6\%$. Note that the large
uncertainty for these measurements is due to high variance for the model
with batch norm, on average $2.3\%$ versus $0.7\%$ for Fixup. JPEG compression
was next at $4.6 \pm 0.3\%$.

Interestingly, some corruptions that led to a
positive gap for VGG8 showed a negative gap for the
WRN, i.e.,~batch norm improved accuracy to:
Contrast---$4.9 \pm 1.1\%$, Snow---$2.8 \pm 0.4\%$, Spatter---$2.3 \pm 0.8\%$.
These were the same corruptions for which VGG13 lost, or did not improve its
robustness when batch norm was removed, hence why we believe these correlate
with standard test accuracy (highest for WRN). Visually, these
corruptions appear to preserve texture information. Conversely,
noise is applied in a spatially global way that disproportionately degrades
these textures, emphasizing shapes and edges.
It is now well known that modern CNNs trained on standard datasets have a
propensity to rely excessively on texture rather than shape
cues~\cite{geirhos2019imagenettrained, brendel2019approximating}.
The WRN obtains $\approx 0$ training error and is in our view over-fitted;
CIFAR-10 is known to be difficult to learn robustly given few
samples~\cite{schmidt2018adversariallya}.

\biblio


\section{Full Results for MLP with Batch Norm}
\label{sec:full-results-mlp}

\begin{table*}[]
\centering
\caption{Mitigating the effect of the input dimension on adversarial
vulnerability by correcting the margin enforced by the loss function.
Regularization constant $\lambda$ is for $\ell_2$ weight decay.
Consistent with~\cite{simon-gabriel2018adversarial}, we use
$\epsilon$-FGSM perturbations, the optimal $\ell_\infty$ attack for a
linear model. Values in rows with
$\sqrt{d} > 28$ are a ratio of entry (accuracy or loss) for the $\sqrt{d} = 28$ baseline. Pred.~is the predicted adversarial damage using Thm.~4
of~\cite{simon-gabriel2018adversarial}, i.e., the increase in
$\| \nabla_x \mathcal{L} \|_{q = 1}$ due to the $\epsilon_\infty$ perturbation.}
\begin{tabular}{cccccccc}
\toprule
\multicolumn{2}{c}{Model} & \multicolumn{2}{c}{Test Accuracy ($\%$) (See Caption)} & \multicolumn{4}{c}{Loss (See Caption)} \\ \midrule
$\sqrt{d}$ & $\lambda$ & Clean & $\epsilon = 0.1$ & Clean & $\epsilon = 0.05$ & $\epsilon = 0.1$ & Pred.\\ \midrule
28 & -- & $92.4 \pm 0.1\%$ & $53.9 \pm 0.3\%$ & $0.268 \pm 0.001$ & $0.646 \pm 0.001$ & $1.410 \pm 0.004$ & -\\ \midrule \midrule
56 & -- & $1.001 \pm 0.001$ & $0.33 \pm 0.03$ & $1.011 \pm 0.007$ & $1.802 \pm 0.006$ & $2.449 \pm 0.009$ & 2 \\ \midrule
56 & 0.01 & $0.999 \pm 0.002$ & $0.98 \pm 0.01$ & $1.010 \pm 0.007$ & $1.010 \pm 0.006$ & $1.01 \pm 0.01$ & - \\ \midrule
84 & -- & $0.998 \pm 0.002$ & $0.10 \pm 0.09$ & $1.06 \pm 0.01$ & $2.84 \pm 0.02$ & $4.15 \pm 0.02$ & 3 \\ \midrule
84 & 0.0225 & $0.996 \pm 0.003$ & $0.96 \pm 0.04$ & $1.05 \pm 0.02$ & $1.06 \pm 0.03$ & $1.06 \pm 0.03$ & - \\ \midrule
112 & -- & $0.992 \pm 0.004$ & $0.1 \pm 0.2$ & $1.18 \pm 0.03$ & $4.15 \pm 0.02$ & $5.96 \pm 0.02$ & 4 \\ \midrule
112 & 0.05 & $0.987 \pm 0.004$ & $1.00 \pm 0.03$ & $1.14 \pm 0.04$ & $1.08 \pm 0.03$ & $1.04 \pm 0.03$ & - \\ \bottomrule
\end{tabular}
\label{tab:linear-model-wd}
\end{table*}

\begin{table*}[]
\centering
\caption{Three-layer ReLU MLP, with and without batch norm (BN), trained for 50 epochs and repeated over five (5) random seeds.
Values in rows with $\sqrt{d} > 28$ are a ratio of entry (accuracy or loss) for the $\sqrt{d} = 28$ baseline.
There is a considerable increase in adversarial vulnerability in terms of the loss, or similarly, a degradation of robustness
in terms of accuracy, due to batch norm. The discrepancy for BIM-$\ell_\infty$ with $\epsilon = 0.1$ for $\sqrt{d} = 84$ with
batch norm represents a $61 \pm 1\%$ degradation in absolute accuracy compared to the baseline.}
\begin{tabular}{cccccccc}
\toprule
\multicolumn{2}{c}{Model} & \multicolumn{2}{c}{Test Accuracy ($\%$) (See Caption)} & \multicolumn{4}{c}{Loss (See Caption)} \\ \midrule
$\sqrt{d}$ & BN & Clean & Noise & $\epsilon = 0.1$ & Clean & $\epsilon = 0.05$ & $\epsilon = 0.1$ \\ \midrule
28 & \xmark & $97.95 \pm 0.08$ & $93.0 \pm 0.4$ & $66.7 \pm 0.9$ & $0.0669 \pm 0.0008$ & $0.285 \pm 0.003$ & $1.06\pm 0.02$\\ \midrule \midrule
28 & \cmark & $0.9992 \pm 0.0012$ & $0.82 \pm 0.01$ & $0.34 \pm 0.03$ & $1.06 \pm 0.04$ & $2.20 \pm 0.03$ & $3.18\pm 0.03$\\ \midrule 
56 & \xmark & $1.0025 \pm 0.0009$ & $1.009 \pm 0.004$ & $0.80 \pm 0.02$ & $0.87 \pm 0.02$ & $1.27 \pm 0.01$ & $1.68 \pm 0.03$\\ \midrule
56 & \cmark & $1.0027 \pm 0.0008$ & $0.853 \pm 0.008$ & $0.13 \pm 0.09$ & $0.91 \pm 0.03$ & $3.48 \pm 0.02$ & $5.83 \pm 0.03$\\ \midrule 
84 & \xmark & $1.0033 \pm 0.0009$ & $1.015 \pm 0.004$ & $0.71 \pm 0.02$ & $0.86 \pm 0.02$ & $1.48 \pm 0.02$ & $2.15 \pm 0.03$\\ \midrule
84 & \cmark & $1.0033 \pm 0.0010$ & $0.865 \pm 0.009$ & $0.09 \pm 0.08$ & $0.88 \pm 0.02$ & $4.34 \pm 0.02$ & $7.34 \pm 0.02$\\ \bottomrule
\end{tabular}
\label{tab:three-layer-bn-full-ratio}
\end{table*}

\biblio


\section{Adversarial and Noisy Examples}
\label{sec:adversarial-examples}

\begin{figure}
\centering
\includegraphics[width=.95\columnwidth]{img/W84_Gauss-u0e+00-std1e+00}
\caption{MNIST samples resized to 84x48 with~\texttt{PIL.Image.NEAREST} and
additive Gaussian noise.}
\label{fig:mnist-d84-noisy}
\end{figure}

We show some upsampled MNIST digits with additive Gaussian noise of the same
mean and variance as the original digits in Figure~\ref{fig:mnist-d84-noisy}.
The noise is much more perceptible than the adversarial examples crafted with
BIM, but the correct digit remains plainly visible to the human eye.